\documentclass[lettersize,journal]{IEEEtran}
\usepackage{amsmath,amsfonts}
\usepackage{algorithm}
\usepackage{array}
\usepackage[caption=false,font=normalsize,labelfont=sf,textfont=sf]{subfig}
\usepackage{textcomp}
\usepackage{stfloats}
\usepackage{url}
\usepackage{cuted}
\usepackage{flushend}
\usepackage{verbatim}
\usepackage{cite}
\usepackage{epsfig}
\usepackage{graphicx}
\usepackage{multirow}
\usepackage{xcolor}
\usepackage{amssymb}
\usepackage{wrapfig}
\usepackage{booktabs}
\usepackage{bm}
\usepackage[font=small]{caption}
\usepackage{algpseudocode}
\newcommand{\bx}{\bm{x}}
\newcommand{\bg}{\bm{g}}
\newcommand{\bp}{\bm{p}}
\newcommand{\bv}{\bm{v}}
\newcommand{\bu}{\bm{u}}
\newcommand{\bX}{\bm{X}}
\newcommand{\bY}{\bm{Y}}
\newcommand{\by}{\bm{y}}
\newcommand{\bV}{\bm{V}}
\newcommand{\bW}{\bm{W}}

\newcommand{\probexp}{\mathbb{E}}
\newcommand{\bepsilon}{\bm{\epsilon}}
\newcommand{\bbias}{\bm{b}}
\newcommand{\bDelta}{\bm{\Delta}}
\newcommand{\bh}{\bm{h}}
\newcommand{\bTheta}{\bm{\Theta}}
\newcommand{\bU}{\bm{U}}
\newcommand{\bLambda}{\bm{\Lambda}}

\newcommand{\abc}[1]{\textcolor{black}{#1}}
\newcommand{\ziquan}[1]{\textcolor{black}{#1}}
\newcommand{\bmu}{\bm{\mu}}
\newcommand{\CUT}[1]{}
\newcommand{\bSigma}{\bm{\Sigma}}
\newcommand{\normdist}{\mathcal{N}}
\def\bnorm{\mathcal{BN}}
\def\Ddistr{\mathcal{D}}
\def\real{\mathbb{R}}
\def\probexp{\mathbb{E}}

\usepackage[pagebackref,breaklinks,colorlinks]{hyperref}

\usepackage[capitalize]{cleveref}
\crefname{section}{Sec.}{Secs.}
\Crefname{section}{Section}{Sections}
\Crefname{table}{Table}{Tables}
\crefname{table}{Tab.}{Tabs.}

\begin{document}

\title{Weight Rescaling: Effective and Robust Regularization for Deep Neural Networks with Batch Normalization}

\author{Ziquan Liu, Yufei Cui, Jia Wan, Yu Mao, Antoni B. Chan\\
Department of Computer Science, City University of Hong Kong

        % <-this % stops a space
%\thanks{This paper was produced by the IEEE Publication Technology Group. They are in Piscataway, NJ.}% <-this % stops a space
%\thanks{Manuscript received May, 2022}
}

% The paper headers
\markboth{Journal of \LaTeX\ Class Files,~Vol.~14, No.~8, August~2021}%
{Shell \MakeLowercase{\textit{et al.}}: A Sample Article Using IEEEtran.cls for IEEE Journals}

%\IEEEpubid{0000--0000/00\$00.00~\copyright~2021 IEEE}
% Remember, if you use this you must call \IEEEpubidadjcol in the second
% column for its text to clear the IEEEpubid mark.

\maketitle
\begin{abstract}
Weight decay is often used to ensure good generalization in the training practice of deep neural networks with batch normalization (BN-DNNs), where some convolution layers are invariant to weight rescaling due to the normalization. In this paper, we demonstrate that the practical usage of weight decay still has some unsolved problems in spite of existing theoretical work on explaining the effect of weight decay in BN-DNNs. On the one hand, when the \ziquan{non-adaptive} learning rate e.g. SGD with momentum is used, the effective learning rate continues to increase even after the initial training stage, which leads to an overfitting effect in many neural architectures. On the other hand, in both SGDM and adaptive learning rate optimizers e.g. Adam, the effect of weight decay on generalization is quite sensitive to the hyperparameter. Thus, finding an optimal weight decay parameter requires extensive parameter searching. To address those weaknesses, we propose to regularize the weight norm using a simple yet effective weight rescaling (WRS) scheme as an alternative to weight decay. WRS controls the weight norm by explicitly rescaling it to the unit norm, which prevents a large increase to the gradient but also ensures a sufficiently large effective learning rate to improve generalization. On a variety of computer vision applications including image classification, object detection, semantic segmentation and crowd counting, we show the effectiveness and robustness of WRS compared with weight decay, implicit weight rescaling (weight standardization) and gradient projection (AdamP). 
\end{abstract}

\section{Introduction}
Because of its benefits for training and generalization, batch normalization (BatchNorm, BN) \cite{ioffe2015batch} has become a standard component in modern neural networks and is widely employed in many machine learning applications, e.g., image recognition \cite{he2016deep}, semantic segmentation \cite{chen2017rethinking} and object detection \cite{redmon2018yolov3}. Despite being widely used, BN is still not well-understood in many aspects. One mystery about deep neural networks with BN (BN-DNNs) is 
\emph{why such a neural network, whose prediction is invariant to scaling of its most weights,} is affected by weight decay (WD), an explicit regularization controlling the $l_2$ norms of weights (denoted as {\em weight norms}). Existing works \cite{zhang2018three,hoffer2018norm,li2020understanding} hypothesize that weight-scale invariant networks, including networks with BN and WeightNorm \cite{salimans2016weight}, need penalties on the weight norms to increase the effective learning rate to achieve faster convergence and better generalization \cite{li2019towards}. In their empirical study \cite{hoffer2018norm,li2020understanding}, the validity of the effective learning rate hypothesis is demonstrated. 
\begin{figure}[t]
  \centering
  \includegraphics[width=1.0\linewidth]{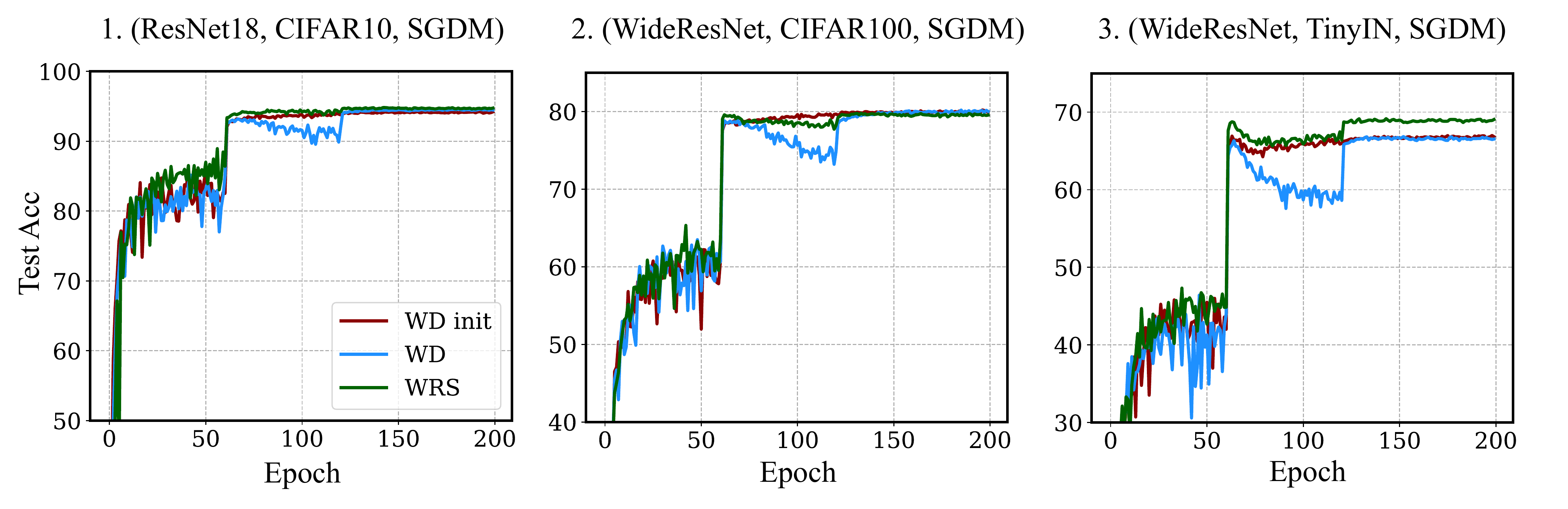}
  \caption{The overfitting caused by weight decay in various neural networks and datasets. The plot titles show the (model, dataset, optimizer).  WD init (weight decay is turned on during 0-60 epochs) has no overfitting effect, while WD (0-200 epochs) overfits after 60 epochs. Since WD increases gradient norms as shown in Fig.~\ref{fig:grad_norm_and_acc}, the experiment indicates that keeping increasing gradient norms leads to overfitting.}
  \label{fig:wd_overfitting}
\end{figure}

\begin{figure*}
  \centering
  \includegraphics[width=1.0\linewidth]{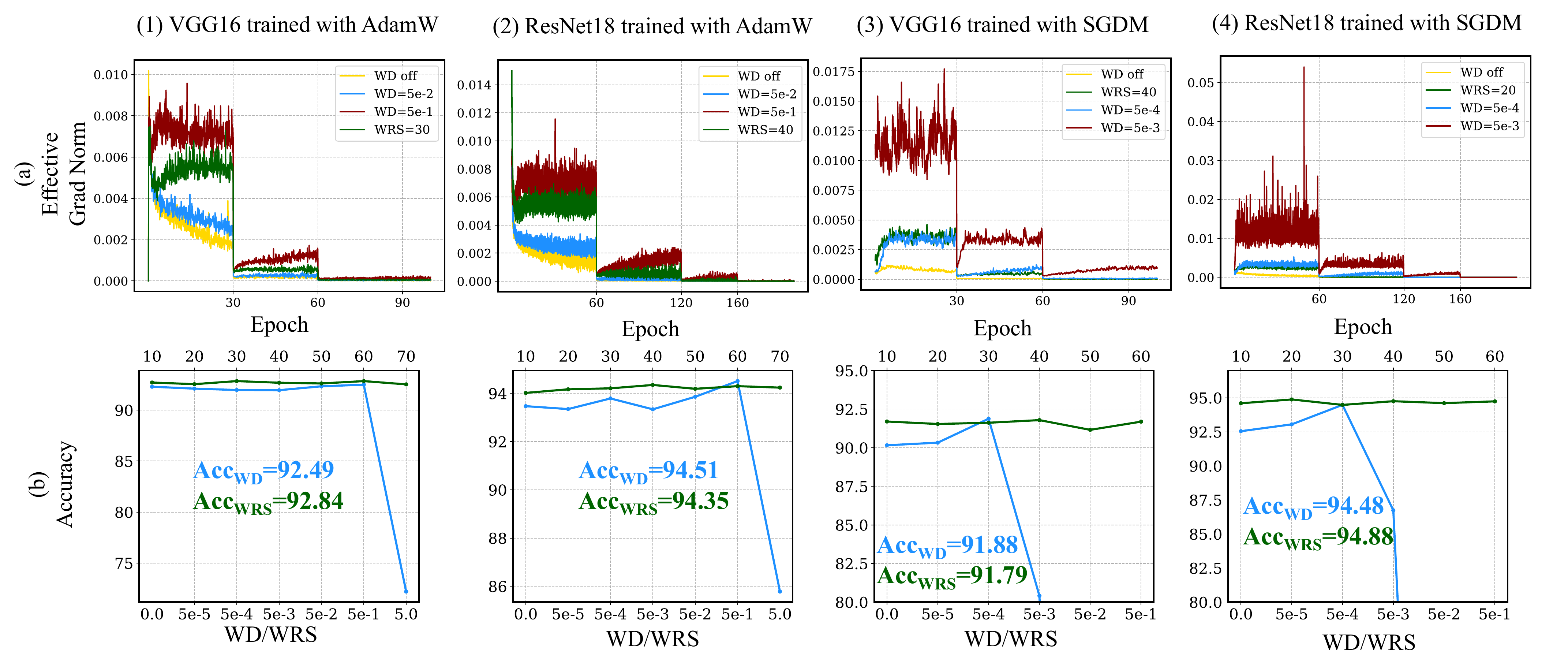}
  \caption{Effective gradient norms and robustness of regularization to hypeparameters. (a) shows the dynamics of effective gradient norms (ratio between gradient norms and weight norms) of an initial layer for VGG16 and ResNet18 trained on CIFAR10. (b) shows the change of accuracy when hyperparameters in WD and WRS are changed, where the top x-axis is the WRS's step parameter and the bottom x-axis is the WD's coefficient. The best test accuracy of WD and WRS is shown with color. In both VGG16 and ResNet18, WRS is less sensitive to the hyperparameter and achieves the better performance than WD, i.e. 92.84 vs. 92.49 in VGG16 and 94.88 vs. 94.51 in ResNet18.}
  \label{fig:grad_norm_and_acc}
\end{figure*}
Although the effective learning rate argument explains the improved generalization of weight decay from a theoretical point of view, the practical usage of weight decay in BN-DNs still has some obstacles. In this paper, we first show the weaknesses of weight decay in optimization with both adaptive and non-adaptive learning rates. 
% and non-adaptive learning rate.
 We use the popular Adam \cite{kingma2014adam} as the representative of adaptive learning rate optimizers, which adjusts learning rates for individual parameters based on their uncentered variance estimates. Non-adaptive learning rate optimizers do not normalize the learning rate based on any moment estimate, e.g. SGD with momentum. For the SGD training with Adam, the effectiveness of weight decay is generally weakened. Figs.~\ref{fig:grad_norm_and_acc}b1 \& b2 show the test accuracy of AdamW \cite{loshchilov2018decoupled} (decoupling weight decay from the Adam momentum estimation) when VGG16-BN \cite{simonyan2014very} and ResNet18 \cite{he2016deep} are trained on CIFAR10 \cite{krizhevsky2009learning} with different hyperparameters of weight decay. The result demonstrates that, even when a large $\lambda$ like 5e-2 is used, the weight decay has almost no effect on the generalization performance of AdamW training. Fig.~\ref{fig:grad_norm_and_acc}a show the gradient magnitude of one conv layer during AdamW training. Compared to no weight decay, the optimization with $\lambda$=5e-2 has the similar gradient norm vanishing phenomenon. For SGD training with a non-adaptive learning rate, Figs.~\ref{fig:grad_norm_and_acc}b3 \& b4 show that the generalization performance is largely determined by the hyperparameters, while Figs.~\ref{fig:grad_norm_and_acc}a3 \& a4 show that the effective learning rate is either too small or too large when an improper hyperparameter is chosen. After the first learning rate decay, the gradient norm still keeps increasing which is shown to lead to an overfitting effect in Fig.~\ref{fig:wd_overfitting}. 

In this paper, we analyze the above weaknesses of weight decay by revisiting the training dynamics of weight in BN-DNNs. We find that to address those difficulties, a proper regularization scheme is needed to both control the weight norm and keep gradient norms stable. Given the above observations, we propose to replace WD by a simple yet effective method that rescales weight norms to the unit norm during training, which we call \emph{weight rescaling (WRS)}. WRS does not affect the network's prediction, but its weight controlling effect prevents the gradient magnitude being too small to escape from bad local minimum and, on the other hand, does not induce a large oscillation of gradient magnitudes. \ziquan{By analyzing the structure of convolutional kernels, it is unveiled that both WRS and WD learn sparse kernels compared with training without regularization, indicating that WRS and WD have similar regularization effects. Finally, the experimental result of tuning hyperparameters on a variety of neural networks and datasets suggests that WRS is not as sensitive as WD to hyperparameters, and hence WRS is a more robust regularization method than WD to hyperparameter variation.}

In summary, our paper has 3 main contributions. First, we revisit the effective learning rate argument of weight decay and unveil the difficulty of using weight decay in optimization with both non-adaptive and adaptive learning rates. Second, we propose to use a simple yet effective regularization method, called weight rescaling (WRS), that controls the weight norm but also prevents large gradient magnitudes from hurting generalization. Finally, WRS is empirically demonstrated to be effective compared to WD, weight standardization \cite{qiao2019weight}, bounded weight normalization \cite{hoffer2018norm} and AdamP \cite{heo2021adamp}, and robust to its hyperparameter across different computer vision tasks, neural architectures and datasets.

The rest of this paper is organized as follows. Section \ref{text:related} discusses related work and Section \ref{sec:network} introduces the background on batch normalization and the empirical difficulties of WD in both adaptive and non-adaptive learning rate optimization. Section \ref{sec:analysisWD} revisits the weight norm dynamics of BN-DNNs, analyzes the reason of WD's weaknesses, and Section \ref{sec:wrs} proposes the weight rescaling scheme to alleviate those problems. Section \ref{text:experiments} gives experimental results in a variety of computer vision applications, which demonstrates the effectiveness of WRS compared with weight decay and several recent baselines.

\begin{figure*}
  \centering
  \includegraphics[width=1.0\linewidth]{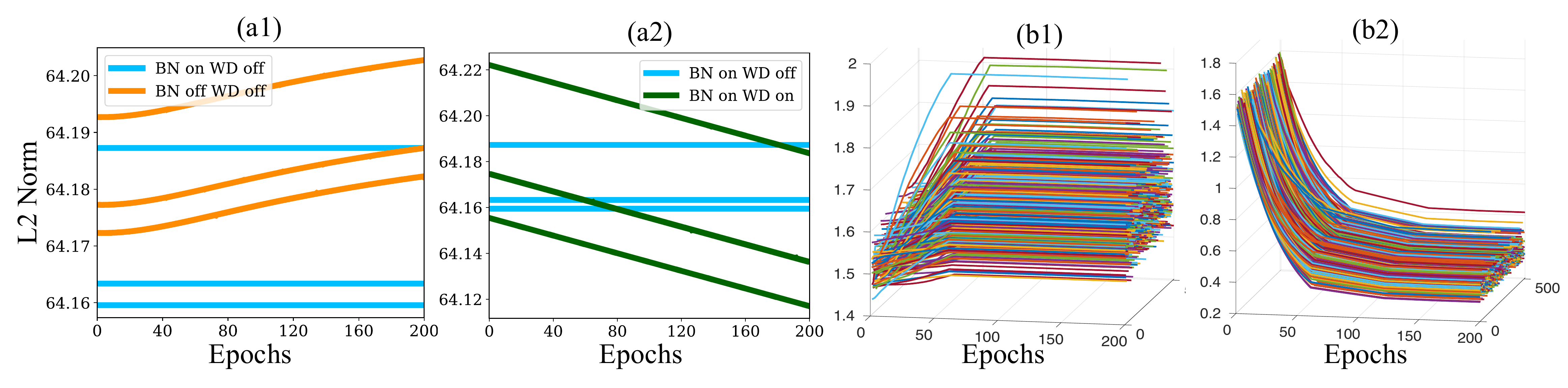}
  \caption{(a) $l_2$ weight norms of MLP during training for BN with and without WD.
  %: BN on (blue), BN off (orange), and BN on and WD on (green). 
  BN without WD forces weight norms to be constant, while using WD decreases weight norms as an exponential function (here appearing locally linear). (b) Weight norms during training of VGG16. Each line denotes the $l_2$ norm of one output channel. (b1) Without WD, weight norms increase at early epochs and remain stable afterwards, when the learning rate is decayed. (b2) With WD, weight norms decrease as an exponential function. The observation also holds in other networks, according to Lemma 3.1.}
  \label{vgg_mnist_l2norm}
\end{figure*}

\section{Related Work}
\label{text:related}

\abc{We first review related work on explaining weight decay regularization with and without BatchNorm. We then review normalization/standardization schemes for improving optimization.}

\subsection{Weight Decay Regularization \abc{and BatchNorm}}
\abc{Several works have studied the effects of weight decay regularization and its effect on BatchNorm DNNs.}
As the first work to investigate the relationship between $l_2$ regularization and normalization in DNNs, \cite{van2017l2} shows that the weight norm is determined by the hyperparameter in $l_2$ regularization and the learning rate, but the relationship between weight norm and generalization is not considered. They also propose to scale weights to the unit norm during training to decouple the entanglement of learning rate and weight decay. However, \cite{van2017l2} is mainly focused on the effective learning rate analysis and does not empirically show the impact of effective learning rate on the performance. Our work extends the analysis of \cite{van2017l2} to practical training and demonstrates the weaknesses of weight decay in terms of optimization. In addition, we carry out extensive experiment evaluations over various tasks and neural networks to show the effectiveness and robustness of WRS. 

\cite{hoffer2018norm, zhang2018three} also explain the effect of WD on the generalization of BN-DNNs via an effective learning rate.\cite{hoffer2018norm} proposes to fix the weight \emph{matrix} norm during training, but their bounded WeightNorm (BWN) is based on WeightNorm, which is shown to be not as good as BatchNorm \cite{gitman2017comparison} in many vision tasks. \cite{liu2020improve} proposes an alternative to WD that avoids the weight-scale-shifting-invariance effect in standard neural networks with ReLU activations, achieving better adversarial robustness than WD. However, the regularization minimizes a product of all weight norms and causes unbalanced weight norms in different layers, which impedes the optimization \cite{neyshabur2015path} and hurts generalization. In contrast, our WRS balances weight norms at different layers and improves generalization. \cite{arora2018theoretical,li2019exponential} consider the convergence rate of BN-DNN with and without WD, while our work focuses on the effect of WD on generalization in practical training. \ziquan{\cite{Roburin2020ASA} analyzes the effect of BN in SGD and demonstrates that the scale-invariant layers make the SGD optimization similar to Adam optimization constrained on a unit-norm hypersphere, while our work handles the problems of WD in practical training and aims to address those issues with WRS.}

\subsection{Weight Norm Constrained Optimization}

\abc{We next review normalization or standardization schemes that aim to improve DNN optimization.}
\cite{huang2017projection} proposes WRS in BN-DNNs but to solve the problem of \emph{weight-scale-shifting invariance} in \emph{deep homogeneous neural networks} \cite{liu2020improve}, i.e., weight scales in different layers can be arbitrarily shifted between layers while keeping the same network function. However, the \emph{scale-invariance} property in BN-DNNs is different from the \emph{scale-shifting invariance} problem in homogeneous DNNs. More importantly, we propose to use WRS as \emph{regularization} to replace WD and demonstrate its effectiveness in various neural architectures and datasets, while \cite{huang2017projection} trains NN with WRS+WD and does not find the regularization effect of WRS. 

Weight standardization (WS) \cite{qiao2019weight} and \cite{huang2017centered} propose to keep weights at unit norm implicitly during training by designing a specific weight function. The fundamental difference is that WS is a reparameterization approach and the normalization function needs to be back-propagated; in contrast, our WRS is independent of the back-propagation. Our experiments show that WS still needs WD to achieve competitive performance (Fig.~\ref{fig:time_effect_and_resnet}b) so the drawbacks of WD is not address in WS. In image classification tasks, when WD is added in WS networks, WRS in large-scale datasets still outperforms WS (Tab.~\ref{table:img_recog}).

\ziquan{AdamP \cite{heo2021adamp} observes that the momentum-based optimizer does not have a large effective learning rate as the momentum will increase the weight norms. To increase the effective learning rate, AdamP proposes to project a weight vector's momentum variable to its orthogonal space so that the weight norm is not rapidly increased. Similar to \cite{heo2021adamp}, \cite{cho2017riemannian} proposes an optimizer on Riemannian manifold, which updates the parameters in tangent space and maps the parameters back to the manifold by parallel translation. In contrast, our work aims to address the instable optimization induced by WD and directly rescales weight norms to prevent weight norms from increasing. Despite its simplicity, we empirically demonstrate that on a large-scale dataset, our rescaling approach is stronger than AdamP. Moreover, when combining AdamP and WRS, the performance can be further improved. Note that we only compare with AdamP since \cite{heo2021adamp} shows that the Grassmann optimizer \cite{cho2017riemannian} does not work as well as AdamP especially on large-scale datasets. }

\ziquan{\cite{zhang2020spherical} proposes to learn embeddings lying on a hypersphere by imposing a regularization term in the objective function, which is the squared $l_2$ distance of the embedding and a pre-defined radius. Our paper aims to regularize the \emph{weights} instead of the last layer's \emph{representations}. The weight norm constraint in our work is explicitly imposed by rescaling instead of minimizing a quadratic regularization function. }

%A classic paper \cite{krogh1992simple} investigates the effect of WD on a one-layer NN by projecting weights onto the input space, and our paper extends \cite{krogh1992simple} to modern BN-DNNs.
% and gives empirical studies on modern NN. 
%\cite{bartlett2020benign} investigates benign overfitting  in linear regression by studying the data covariance matrix, which is similar to our analysis on the input feature covariance matrix. However, our focus is on  more complex DNNs. 
%%%%%%%%
\CUT{
Finally, our analysis on the input feature span is closely related to kernel methods \cite{belkin2018understand}. In kernel methods, SGD/GD initialized outside the span of kernel features cannot converge to the minimum norm solution since the components orthogonal to the input space are not updated.  We show that SGD for a BN-DNN without regularizing on the components orthogonal to the input feature span cannot converge to a solution with succinct weights.
}
%%%%%%%%

\section{Background and Issues of Weight Decay}
\label{sec:network}

%In this section 
We first introduce notations of a BN-DNN and then demonstrate the practical issues of weight decay in this section. The notation and analysis are based on a fully-connected network, and a similar analysis on convolution neural networks is shown in the Appendix.
\subsection{BatchNorm Deep Neural Network (BN-DNN)}
 \label{sec:BN-DNN}
A BN-DNN is denoted as $f_{\bTheta}: \mathbb{R}^D \mapsto \mathbb{R}^K$, where $D$ is the dimension of an input sample $\bx_i$ and $K$ is the number of classes in classification or the dimension of output in regression. The input $\bX=\{\bx_i\}_{i=1}^N$ and label $\bY=\{\by_i\}_{i=1}^N$ are sampled from an unknown distribution $\mathcal{D}$. We assume that the BN-DNN is composed of a number of BN layers, comprising a sequence of weight multiplication, batch normalization and activation function \cite{ioffe2015batch}. The BN layer is %formulated as
\begin{align}
\bh_{l+1}^{(i)}=\phi(\mathcal{BN}(\bW_{l}^T\bh_{l}^{(i)})),
\end{align}
where $\bh_{l}^{(i)}\in\mathbb{R}^{H_{l}\times 1}$ is the $i$th hidden variable of $l$th layer,  $\bW_{l}\in\mathbb{R}^{H_{l}\times H_{l+1}}$ is the weight matrix of $l$th layer, $\bnorm$ represents the BatchNorm operation, $\phi(x)$ is the ReLU activation function, and we define the input as $\bh_0^{(i)}=\bx_i$. The BatchNorm operation is an element-wise operation,
\begin{align}
\hat h_{l+1,j}^{(i)}&=\bnorm(\bW_{l, j}^T\bh_{l}^{(i)})=\gamma_{l,j} t_{l,j}^{(i)} +\beta_{l,j}, \nonumber\\ 
t_{l,j}^{(i)} &= \frac{\bW_{l,j}^T\bh_{l}^{(i)}-\bW_{l,j}^T\bmu_{l}^{(B)}}{\|\bW_{l,j}\|_{\bSigma_{l}^{(B)}}}, 
\label{equ:bn}
\end{align}
where $t_{l,j}^{(i)}$ denotes the normalized output, and 
$\|\bW_{l,j}\|_{\bSigma_{l}^{(B)}}=(\bW_{l,j}^T\bSigma_{l}^{(B)}\bW_{l,j})^{1/2}$ is the standard deviation of this neuron's output.
% and we use $t_{l,j}^{(i)}$ to denote the normalized output, i.e. $\hat h_{l+1,j}^{(i)}=\gamma_{l,j}t_{l,j}^{(i)}+\beta_{l,j}$. 
$\bmu_{l}^{(B)}$ and $\bSigma_{l}^{(B)}$ are the mean and covariance matrix of the  batch, 
\begin{align}
\bmu_{l}^{(B)}&=\frac{1}{B}\sum\nolimits_{i=1}^B\bh_{l}^{(i)},\nonumber\\ 
\bSigma_{l}^{(B)}&=\mathrm{ddiag}\big(\frac{1}{B}\sum\nolimits_{i=1}^B\bh_{l}^{(i)}{\bh_{l}^{(i)}}^T-\bmu_{l}^{(B)} {\bmu_{l}^{(B)}}^T\big),\nonumber
\end{align}
%\abc{where $\mathrm{ddiag}(\mathbf{A}) = \mathbf{A}\circ \mathbf{I}$ keeps the diagonal of matrix $\mathbf{A}$ (zeroes out the off-diagonal elements), and $\circ$ is the element-wise product.}
\ziquan{Note that the covariance $\bSigma_{l}^{(B)}$ is written in this form for simplicity but the actual computation is efficient \ziquan{since the variance is computed after the weight multiplication $\bW_{l, j}^T\bh_{l}^{(i)}$}, i.e., the full covariance matrix is not computed.} Let there be $L$ BN layers and one output layer in the network, i.e. $\bTheta=\{\bW_0,\dots, \bW_{L}\}$ where $\bW_{L}\in\mathbb{R}^{H_L\times K}$. The network is trained with mini-batch SGD to minimize the empirical risk $\hat R$ (e.g., cross-entropy or mean-squared error) over dataset
% $\{\bx_i,\by_i\}_{i=1}^N$
$\{\bX,\bY\}$,  %where $R$ can be cross-entropy or mean squared error, 
\begin{align}
\bTheta^*=\min_{\bTheta}{\hat R}_{\bTheta}(\bX,\bY)=\min_{\bTheta}\frac{1}{N}\sum\nolimits_{i=1}^{N}r^{(i)},
\end{align}
where $r^{(i)}$ is shorthand for $r(f_{\bTheta}(\bx_i),\by_i)$. The $l_2$ norm of a vector is denoted as $\|\cdot\|_2$ and the Frobenius norm of a matrix is denoted as $\|\cdot\|_F$. \ziquan{In this paper, we mainly consider the classifier prediction, so the argmax prediction is invariant to weight rescaling though the output layer is not invariant to rescaling the matrix $\bW_L$.}

\CUT{It is obvious from Equation (\ref{equ:bn}) that the output of a BN layer is invariant to the weight norm scaling. \cite{van2017l2} argues that the effect of WD is mainly controlling the effective learning rate and assumes that the reciprocal of gradient norm has a linear relationship with the weight norm. Here we derive the gradient of weight to show that the reciprocal of gradient norm actually scales quadratically with the weight norm. The following is the gradient of $\bW_{l,j}$
\begin{align}
\frac{\partial R_{\bTheta(t)}(\bX^{(B)},\bY^{(B)})}{\partial \bW_{l,j}(t)}=\sum_{i=1}^B\frac{\partial r^{(i)}(t)}{\partial \hat h_{l+1,j}^{(i)}(t)} \frac{\partial \hat h_{l+1,j}^{(i)}(t)}{\partial \bW_{l,j}(t)},\nonumber
\end{align}
\begin{align}
\gamma_{l,j}\left[\frac{\bh_l^{(i)}-\bmu_l^{(B)}}{\|\bW_{l,j}\|_{\bSigma_{l}^{(B)}}}  - \frac{\bW_{l,j}^T(\bh_{l}^{(i)}-\bmu_{l}^{(B)})}{   \|\bW_{l,j}\|^3_{\bSigma_{l}^{(B)}}  } \bSigma_{l}^{(B)}\bW_{l,j}\right],
\end{align}}

\subsection{Issues of Weight Decay}
\label{sec:issure_wd}
Although the prediction and generalization of BN-DNNs are invariant to weight norms, using weight decay to control the weight norms during SGD training is still one of the most popular techniques to improve the generalization. The common \abc{hypothesis to explain} %opinion on 
this phenomenon is that decreasing the weight norm in effect magnifies the learning rate, which is needed for SGD optimization to achieve good generalization. However, the theoretical explanation does not settle the issues of WD in practical training. Here we reveal the weaknesses of weight decay training in Adam and SGD (with momentum) respectively. 

\paragraph{Adaptive learning rate} To demonstrate the problem with weight decay in optimization with adaptive learning rate e.g. AdamW, we train two representative networks (VGG16-BN\footnote{We add BatchNorm to all conv layers in VGG16 throughout this paper.} \cite{simonyan2014very} and ResNet18 \cite{he2016deep}) on CIFAR10 \cite{krizhevsky2009learning} using AdamW. In VGG16, we set batch size as 100, total training epoch as 100 and initial learning rate as 0.001. The learning rate is divided by 10 in the 30th and 60th epochs. In ResNet18, we set batch size as 100, total training epoch as 200 and initial learning rate as 0.001. The learning rate is divided by 10 in the 60th, 120th and 160th epochs. In Fig.~\ref{fig:grad_norm_and_acc}a we visualize dynamics of effective gradient norms of a shallow layer (VGG16's 2nd layer and ResNet18's 3rd layer), which is defined as the ratio of gradient norm and weight norm. The accuracy for different hyperparameters of WD is visualized in Fig.~\ref{fig:grad_norm_and_acc}b. The gradient norm is computed using the normalized gradients by the second-order momentum, i.e. the actual gradients applied to parameters.

There are several main observations in the experiment. First, comparing the effective gradient norm of WD off and WD=5e-2 (Fig.~\ref{fig:grad_norm_and_acc}a1 and a2), the effective gradient norm in Adam training is quite insensitive to the weight decay hyperparameter within some range. Only when a relatively large WD is used (5e-1) will the effective learning rate be obviously increased. Second, WD improves the generalization in a small range of hyperparameters. For example, in Fig.~\ref{fig:grad_norm_and_acc}b2 only WD=5e-1 improves the generalization by an obvious margin. Third, when a good hyperparameter of WD is used, \ziquan{i.e., WD=5e-1}, the effective gradient norms continues to increase as a result of decreasing weight norms, especially in the middle stage of optimization, which may lead to unstable optimization and performance degradation.
% \ziquan{(Put the summary in the end of this section.)}

\paragraph{Non-Adaptive learning rate} The non-adaptive learning rate experiment setting is the same as adaptive learning rate one, except that the initial learning rate is 0.1. To better visualize the trend of effective gradient norms, we compute the exponential moving average of gradient norms with 0.6 as the exponent in Fig.~\ref{fig:grad_norm_and_acc}a3 and a4. First, the effective gradient norm is sensitive to the WD hyperparameter, i.e. WD=5e-4 is good enough for improved generalization. However, the sensitivity also leads to the small range of good hyperparameters: when WD=5e-3 the effective gradient norm is too large and the resulting generalization is severely affected. Second, we also observe that the effective gradient norm consistently increases (even after the learning rate decay) as a result of consistently regularized weight norms. Fig.~\ref{fig:wd_overfitting} shows the ablation study that indicates the regularization effect results in an overfitting effect on three image classification datasets.

\ziquan{In conclusion, we unveil the instability of gradient as training proceeds when a large weight decay is used in both optimizers and the difficulty of hyperparameter selection as a result of sensitivity to WD parameter. In the following sections, we first introduce the weight norm dynamics in BN-DNN training and provide our explanation for the observed phenomenon.}

\section{Analysis of Weight Decay}
\label{sec:analysisWD}
We next introduce the dynamics of weight norms during training and give an intuition on the reason why WD is not robust in practical training. \abc{We consider both the ideal case of continuous-time optimization with infinitesimal learning rate, as well as the practical case of discrete-time optimization with larger learning rate.} Based on the analysis, we propose to use an explicit weight norm control to regularize BN-DNNs and show its effectiveness and robustness with empirical evidence.

\label{sec:sgd_low_rank}
\subsection{Dynamics of Weight Norms in Continuous-time Domain}
Gradient flow, i.e., SGD with an infinitesimal learning rate, provides insights on the optimization of DNNs \cite{du2018gradient, arora2018optimization}. Here we revisit the analysis on dynamics of weight norms during training via gradient flow \cite{arora2018theoretical}. Using the chain rule, the gradient flow of the $l_2$ norm of weight vector $\bW_{l,j}(t)$ is
\begin{align}
\tfrac{d\|\bW_{l,j}(t)\|_2^2}{dt}=2\langle \bW_{l,j}(t), \tfrac{d\bW_{l,j}(t)}{dt} \rangle,
\label{diff_bn}
\end{align}
where the gradient flow of $\bW_{l,j}(t)$ follows the negative direction of the gradient of loss since we use SGD. By deriving the gradient of loss w.r.t. $\bW_{l,j}(t)$, we have the following lemma \cite{arora2018theoretical} (complete proofs appear in Appendix A).

\vspace{0.2cm}
\noindent\textbf{Lemma 3.1} \emph{Given a network as in Sec.~\ref{sec:BN-DNN}, which is trained using SGD to minimize the loss $\hat R_{\bTheta}$ (\abc{i.e., without weight decay}), the gradient flow of the weight norms in BN layers is zero, i.e., 
$\frac{d\|\bW_{l,j}(t)\|_2^2}{dt}=0.$}
%\begin{align}
%frac{d\|\bW_{l,j}(t)\|_2^2}{dt}=0.
%\label{Thm3}
%\end{align}}
\vspace{0.2cm}

There are several implications \abc{of not using WD} arising from Lemma 3.1. First, when the learning rate is infinitesimal, the weight norms in BN layers remain the same as their initializations. Second, in the continuous-time domain, the weight in a BN layer only changes its direction and the scaling is controlled by another parameter $\gamma_{l,j}$, which is the same as WeightNorm \cite{salimans2016weight}.  \ziquan{Figure \ref{vgg_mnist_l2norm}a-b shows the dynamics of weight norm in MLP and VGG16. To simulate the infinitesimal learning rate, we set the learning rate as 1e-6. Thus, the behavior of weight norm in Figure \ref{vgg_mnist_l2norm}a is quite different from the large learning rate training in other figures. }

\abc{Next we consider the weight norm during optimization when weight decay is used, yielding the following corollary.}

\vspace{0.2cm}
\noindent\textbf{Corollary 3.2} \emph{Given a network as in Sec.~\ref{sec:BN-DNN}, which is trained using gradient descent to minimize the loss $\hat R_{\bTheta}$ and $l_2$ regularization $\frac{\lambda}{2}\sum_{l=0}^{L}\|\bW_l\|_F^2$, the weight norm in a BN layer follows an exponential decay, i.e., 
$\|\bW_{l,j}(t)\|_2^2=\|\bW_{l,j}(0)\|_2^2\exp{(-2\lambda t)}$.}
%\begin{align}
%\|\bW_{l,j}(t)\|_2^2=\|\bW_{l,j}(0)\|_2^2\exp{(-2\lambda t)}.
%\end{align}}
\vspace{0.2cm}

%See the Supp.~A.2 for the proof. 
Hence using WD in this domain leads to an exponential decay of  weight norm.
Fig.~\ref{vgg_mnist_l2norm}a shows an example of decaying weight norms due to WD in fully-connected NNs. The exponential decay of WD indicates that the gradient flow of SGD with WD leads to shrinking weight norms, and the optimal solution is weights with zero norm. Thus, the dynamics lead to a degenerate solution, and may cause poor optimization procedures. To address this, we propose a better training scheme, WRS, which has the same regularizing effect and does not lead to degenerate solutions.

\subsection{Dynamics of Weight Norms in Discrete-time Domain}
The infinitesimal learning rate assumption in the continuous domain is not realistic since in practice the initial learning rate is generally large \cite{zhang2018three}. Fortunately, given the orthogonality between gradients and weights, it is straightforward to derive a discrete-time weight norm dynamics. The change of norm is related to the difference between the weight norms of two consecutive optimization steps, i.e. $\bDelta_{l,j}(t)=\|\bW_{l,j}(t+1)\|_2^2-\|\bW_{l,j}(t)\|_2^2$. Let the learning rate be $\eta(t)$, then we have the following corollary \abc{when WD is not used}.

\vspace{0.2cm}
\noindent\textbf{Corollary 3.3} \emph{Given network as in Sec.~\ref{sec:BN-DNN} and the network is trained by SGD to minimize $\hat R_{\bTheta}(\bX,\bY)$ \abc{(i.e., without weight deay)}, we have
$\bDelta_{l,j}(t)=\eta(t)^2\|\frac{\partial \hat R_{\bTheta(t)}(\bX^{(B)},\bY^{(B)})}{\partial \bW_{l,j}(t)}\|_2^2.$}
\vspace{0.2cm}
%\begin{align}
%\bDelta_{l,j}(t)=\eta(t)^2\|\frac{\partial R_{\bTheta(t)}(\bX^{(B)},\bY^{(B)})}{\partial \bW_{l,j}(t)}\|_2^2.
%\end{align}}

When the learning rate $\eta(t)$ goes to infinitesimal, the training enters the continuous-time domain and Lemma 3.1 holds, i.e. $\bDelta_{l,j}(t)\rightarrow 0$. The corollary shows that in practice, \abc{without weight decay,} the weight norm always increases and the amount of increase is determined by the magnitude of its gradient and the learning rate.
%This corollary validates the intuition that weight norms will go unbounded in a BN-DNN
%without WD \cite{hoffer2018norm,van2017l2}, and provides a quantitative increasing rate for weight norms. 
Fig.~\ref{vgg_mnist_l2norm}b shows the change of norms in one convolution layer of VGG16 \cite{simonyan2014very}.
%, where each line denotes a conv kernel of one output channel. 
%This empirical result together with our analyses in Sec.~\ref{sec:sgd_low_rank} demonstrates that there will be a large accumulated noise in weights because of random initialization and noise in SGD at the initial training stage, if WD is not used.

As mentioned in previous sections, weight norms have no effect on the prediction and thus generalization of a BN-DNN. However, the weight norm is related to the gradient norm or effective learning rate in existing literature \cite{hoffer2018norm}. Specifically, the gradient norm is proportional to the reciprocal of weight norm.
\begin{align}
\|\nabla \bW_{l,j}\| \propto \frac{1}{\|\bW_{l,j}\|_{\bSigma_{l}^{(B)}}} \propto \frac{1}{\|\bW_{l,j}\|_2}.
\end{align}
The weight norm dynamics indicate that if there is no weight decay or other forms of weight constraint, weight norms in BN-DNNs will continuously increase so the gradient norm will continuously decrease (see yellow lines in Fig.~\ref{fig:grad_norm_and_acc}a). The decreased gradient norms lead to unsatisfactory generalization and slow convergence during training \cite{li2019towards}. Therefore, during SGD optimization of a BN-DNN, a weight controlling scheme is necessary to maintain sufficiently large gradients for better convergence and generalization. However, Section \ref{sec:issure_wd} reveals the problems of weight decay when used in practical training. We next analyze the reason for those issues by investigating the effective learning rate of WD in Adam and SGD.

\begin{figure*}
  \centering
%  {\small \hspace{0.3cm}(a)  input space projection \hspace{1cm} (b) original space}
  \includegraphics[width=1.0\linewidth]{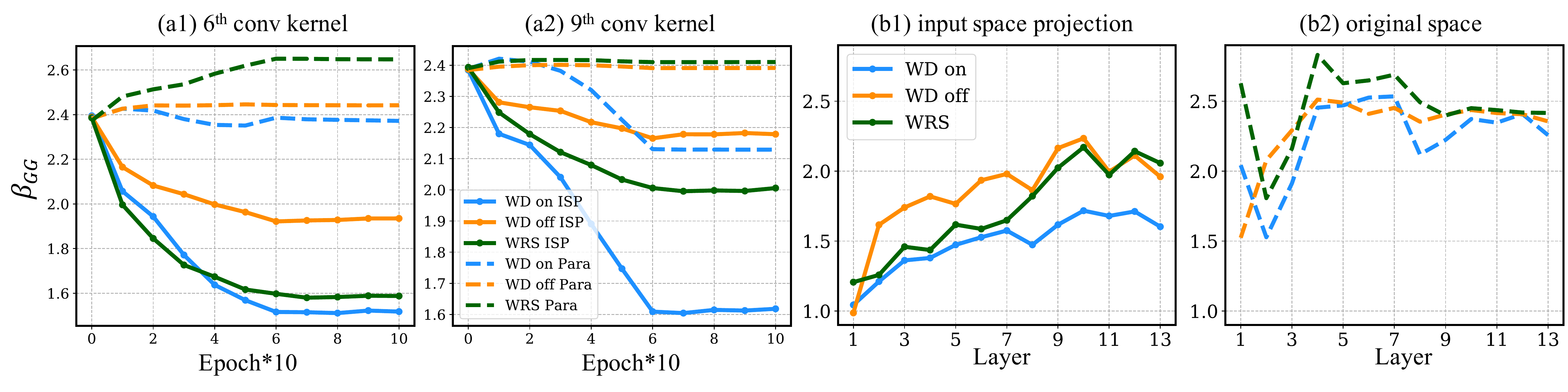}
  \caption{ (a) Shape parameters $\beta_{GG}$ of generalized Gaussians vs.~training epochs for the 6th and 9th conv kernels. Dashed lines and solid lines denote the GGD for conv kernels in parameter space (Para) and input space projection (ISP). The weight sparsity changes dramatically at early training epochs, and remains stable after 60 epochs because of learning rate decay. (b) Shape parameter $\beta_{GG}$ of generalized Gaussians for weights in (b1) input space projection (ISP) and (b2) original space, versus conv layer. WD and WRS lead to more sparse weights at early layers in the ISP instead of the original space, compared to without WD.
  }
  \label{fig:beta_fig}
\end{figure*}

%In this section, we introduce two issues of weight decay in the practical training of BN-DNNs with SGD and Adam optimizers. 

\subsection{The Effective Learning Rate in Adam}
In the previous section, the effective learning rate is shown to be increasing as the weight norm is decreasing. However, in the adaptive learning rate of Adam, the effect of increased effective learning rate is reduced. Recall that the Adam optimizer has the following update rule,
\begin{align}
\nabla \theta_t &:= \nabla_{\theta}\hat R(\theta),\\
m_t &:= \beta_1m_{t-1}+(1-\beta_1)\nabla \theta_t,\\
v_t &:= \beta_2v_{t-1}+(1-\beta_2)\nabla \theta_t^2,\\
\hat m^t &:= m_t/(1-\beta_1^t),\\
\hat v^t &:= v_t/(1-\beta_2^t),\\
\theta_t &:= \theta_{t-1} -\eta \hat m_t /(\sqrt{\hat v_t}+\epsilon). 
\end{align}
where the $\beta_1$ and $\beta_2$ \abc{momentum values} are often set as 0.9 and 0.999. Compared with SGD+momentum, the actual learning rate in Adam is $\eta \hat m_t /(\sqrt{\hat v_t}+\epsilon)$ which is adaptive based on the uncentered second-order momentum estimate and related to the gradient norms. Recall that the gradient norm in BN is proportional to the reciprocal of weight norm, which makes the effective learning rate invariant to weight norm. Thus in Figs.~\ref{fig:grad_norm_and_acc}a1 \& a2, even when weight decay is applied, the effective gradient norm still decreases and the performance improvement is not evident. Only when a large $\lambda$ is used, the effective gradient norm can be increased as a result of decreased weight norm. On the other hand, as indicated by Figs.~\ref{fig:grad_norm_and_acc}b1 and b2, when $\lambda$ is too large, the resulting generalization will deteriorate as a result of oscillating gradients. \ziquan{Thus, selecting an optimal WD parameter requires a non-trivial hyperparameter search as a result of the vulnerability of training to hyperparameter selection. }

\begin{figure}
  \centering
  \includegraphics[width=1.0\linewidth]{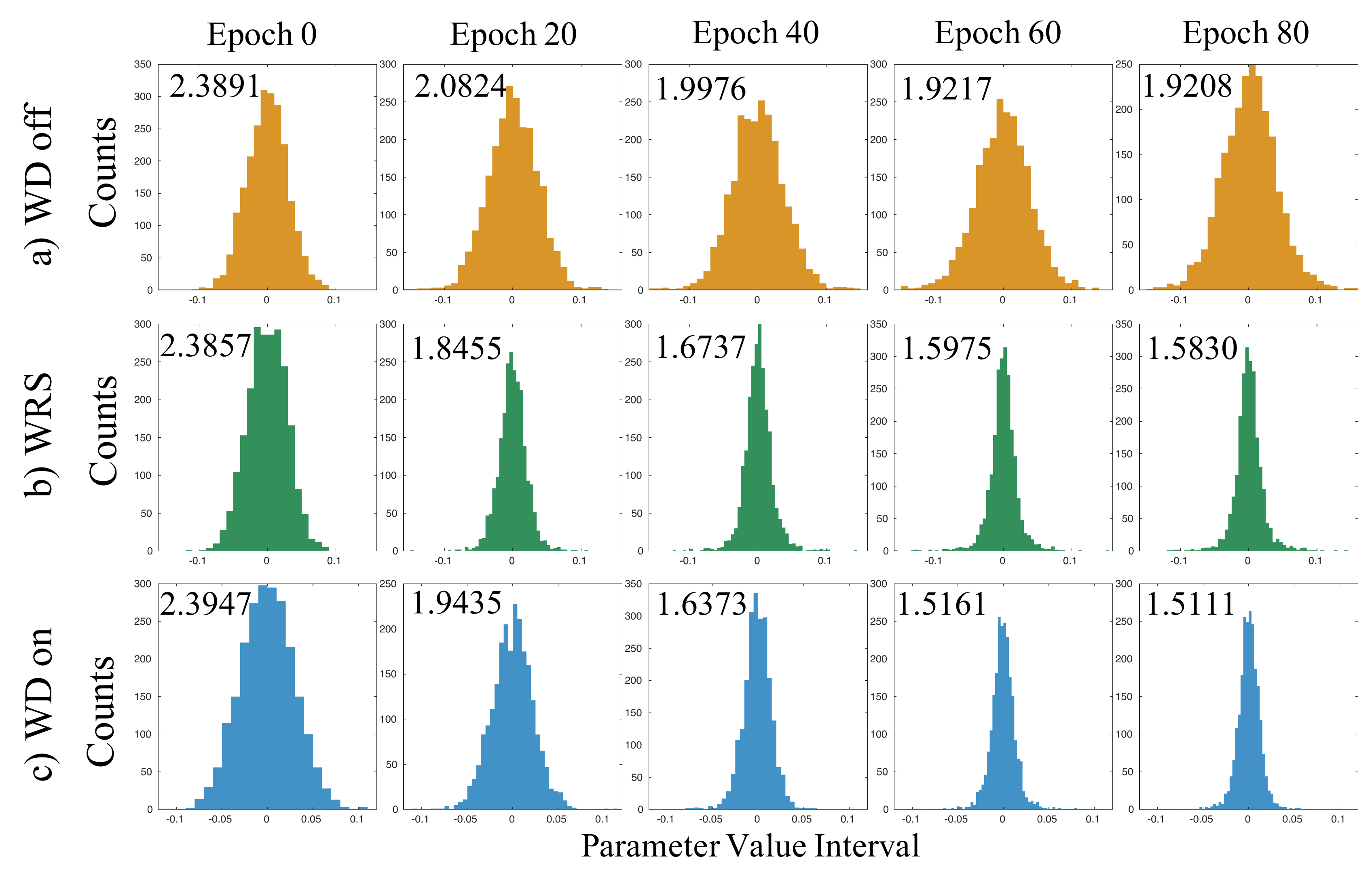}
  \caption{Change of parameter distributions during training after input span projection of the 6th conv parameters, when training with a) WD off, b) WRS and c) WD. The number in each plot denotes the shape parameter $\beta_{GG}$ of the fit generalized Gaussian.
  }
  \label{fig:chang_hist}
\end{figure}

\subsection{The Instable Optimization Caused by WD}
When a non-adaptive learning rate is used, the effective gradient norm increases drastically as weight decay is enforced. So the optimal $\lambda$ in SGDM is much smaller than that in AdamW. At the same time, because of the sensitivity of SGDM to $\lambda$, the optimal value also needs careful tuning in practice. Another interesting finding is that during the middle stage of training, the effective gradient norm increases as a result of decreased weight norms as shown by Figs.~\ref{fig:grad_norm_and_acc}a3 \& \ref{fig:grad_norm_and_acc}a4. To investigate how the increasing weight norm affects the performance, we visualize the test accuracy of ResNet18 trained on CIFAR10 when an optimal $\lambda$ is used for all epochs and for only 0-60 epochs in Fig.~\ref{fig:wd_overfitting}. The result demonstrates that during the middle training stage, weight decay makes the network to overfit as a result of large gradient norms. The same phenomenon can also be observed in a larger network (WideResNet \cite{Zagoruyko2016WRN}) and larger datasets (CIFAR100 and Tiny ImageNet \cite{deng2009imagenet}) in Fig.~\ref{fig:wd_overfitting}.  This overfitting effect is related to the time effect of weight decay \cite{golatkar2019time}, which means that the weight decay mainly works at the initial training stage and has virtually no effect if applied after the learning rate decay. Our work connects the overfitting effect of WD after the learning rate decay to the consistently increasing gradient norms.

\section{Weight Rescaling Regularization}
\label{sec:wrs}
The analyses and observations from the previous sections brings a question about the regularization effect of weight decay: can we have a regularization method that constrains the weight norm but does not lead to drastically large gradients and unsatisfactory training behavior in AdamW and SGD? One simple solution is to only apply weight decay at the initial training stage, i.e. before the learning rate decay. However, such a solution does not solve the issue of hyperparameter sensitivity of weight decay. 

In this paper, we consider a simple but effective solution to this problem, i.e. rescaling weight norm to the unit norm every $\tau$ steps of SGD update, which we denote as weight rescaling (WRS).  WRS has several benefits in practice. First, the hyperparameter $\tau$ in WRS does not need careful tuning. Fig.~\ref{fig:grad_norm_and_acc}b shows the robustness of performance with respect to the step parameter in WRS compared with the sensitivity of WD.  Second, the gradients in WRS are more stable than those in WD as shown by Fig.~\ref{fig:grad_norm_and_acc}a. The stable gradients do not incur the overfitting effect after the learning rate decay as shown by Fig.~\ref{fig:wd_overfitting}.  Finally, the implementation of WRS is easy and there is almost no additional time cost when WRS is used. In summary, the WRS is an effective and robust alternative to WD in practical training. In the experiment, we give more results of WRS on more computer vision tasks, including training from scratch and fine-tuning from a pre-trained model. \ziquan{We present the algorithm of WRS in Algorithm~\ref{alg:wrs}. Note that the weight parameters in convolution or fully connected layers are not included in the WD,  while the BN layers' parameters are penalized by WD.}

\subsection{The Regularization Effect of WRS on Model Parameters}
\ziquan{To observe how WRS regularizes the weights, we investigate the distribution of learned convolution kernel parameters for VGG16. We fit a generalized Gaussian distribution (GGD) \cite{nadarajah2005generalized} to the learned weight values in a convolution layer, \abc{where we use all values in the weight matrix to} %. Specifically, the weight parameters are considered as samples from a distribution and we use all values in the weight to
 estimate the GGD's parameters. The shape parameter $\beta_{GG}$ of the GGD controls the tail shape of the distribution: $\beta_{GG}=2$ corresponds to a Gaussian distribution; $\beta_{GG}<2$ corresponds to a heavy-tail distribution with lower values giving fatter tails ($\beta_{GG}=1$ is a Laplacian); $\beta_{GG}>2$ gives light-tail distributions, and $\beta_{GG}\rightarrow\infty$ yields uniform distribution around 0 (i.e., no tails). In other words, a large $\beta_{GG}$ indicates that more kernel parameters are around zero, while a small $\beta_{GG}<2$ indicates the kernel has a more sparse structure. }

\ziquan{Using a well-trained VGG16 on CIFAR10, we fit GGDs for kernel parameters in all 13 conv layers, and plot the $\beta_{GG}$ versus layers in Fig.~\ref{fig:beta_fig}b before and after projecting weights to their input span. The input space projection is computed by collecting features of all training data in each layer and computing $\bU_l$ by eigendecomposition of $\hat\bSigma_l$, which is estimated by
%%. The $\hat\bSigma_l$ is estimated by
\begin{align}
    \hat\bSigma_l = \frac{1}{N_{tr}}\sum_{i}^{N_{tr}}\bh_l^{(i)}{\bh_l^{(i)}}^T-\bmu_l\bmu_l^T.
\end{align}
Fig.~\ref{fig:beta_fig}b has three implications. First, from the first layer to the final layer, conv parameters become less sparse if we observe after the input span projection (ISP). Second, when we consider ISP of weights, SGD with WD leads to more sparse kernels than vanilla SGD (Fig.~\ref{fig:beta_fig}.b1). However, the sparsity cannot be observed from the parameter space directly (Fig.~\ref{fig:beta_fig}.b2). This result demonstrates that it is necessary to check the regularization effect on sparsity from the ISP space and we next compare the WD and WRS in inducing sparse kernels. }

\begin{figure*}
  \centering
%  {\footnotesize \hspace{0.5cm} (a) with WRS \hspace{2.3cm} (b) with WD}
  \includegraphics[width=1.0\linewidth]{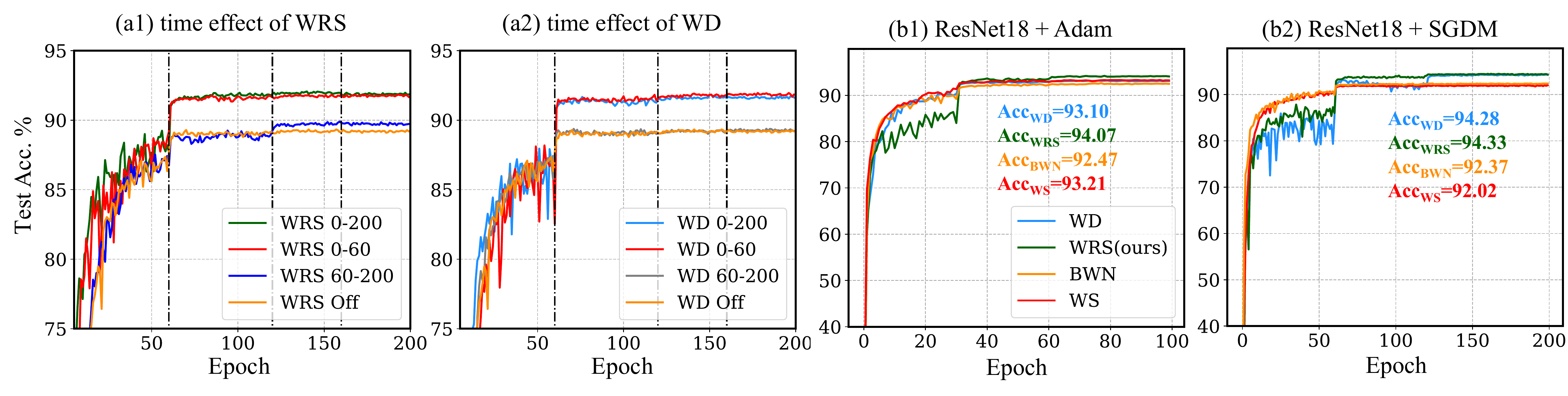}
  \caption{(a) Training curves when training VGG16 on CIFAR10 using WRS and WD in different epochs (0-200, 0-60, 60-200). Black dashed lines denote epochs at which the learning rate is decayed by 0.1. The effect on generalization of both WRS and WD is time-dependent. (b) Learning curves of WRS, Bounded WeightNorm (BWN) \cite{hoffer2018norm}, WD and Weight Standardization (WS) \cite{qiao2019weight} on ResNet18-variant and CIFAR10. WRS has the best performance even though it does not achieve the fastest convergence at early epochs. WS without WD cannot achieve a competitive performance, indicating the fundamental difference between WRS and WS. }
  \label{fig:time_effect_and_resnet}
\end{figure*}

\ziquan{Fig.~\ref{fig:beta_fig}a shows that when observed from ISP, in the 6th conv layer, WRS has a similar distribution shape to WD, while in the deeper 9th layer, WRS's distribution shape is between WD and vanilla SGD. However, if we observe from the parameter space, WRS has the least sparse weight structures in both layers. In Fig.~\ref{fig:chang_hist}, the histograms of conv parameters after ISP show that training with WRS or WD leads to sparse distributions, but without WD the distribution looks similar to a Gaussian. Fig.~\ref{fig:beta_fig}b further demonstrates that WRS has similar sparse structures as WD in initial layers, and similar parameter distributions as WD-off training at final layers, indicating the network needs a sparse structure at initial layers instead of final layers to achieve better performance. We also visualize the time-dependent effect of WD and WRS on VGG16 and CIFAR10 in Fig.~\ref{fig:time_effect_and_resnet}. Same as WD, WRS has its main effect during the initial training stage. The difference is that WRS improves the generalization even when applied in the final training stage.}

\renewcommand{\algorithmicrequire}{\textbf{Input:}}
\renewcommand{\algorithmicensure}{\textbf{Output:}}
\begin{algorithm}
\caption{The Algorithm of Weight ReScaling (WRS)}\label{alg:cap}
\begin{algorithmic}[1]
\Require Training data $\Ddistr_{tr}$, Initialized model $f_{\bTheta}$, Empirical Risk $r(\cdot,\cdot)$, Maximum Step $T$, Batch Size $B$, Learning Rate $\eta$, Weight Decay $\lambda$, WRS Step $\tau$.
\Ensure Optimized model $f_{\hat\bTheta}$.
\State $t\gets 1$.
\State Define $R(\{\bx_i,y_i\}_{i=1}^B)=\frac{1}{B}\sum_{i=1}^B\frac{\partial r(f_{\bTheta}(\bx_i),y_i)}{\partial \bTheta}$.
%\State $X \gets x$
%\State $N \gets n$
\While{$t < T$}
\State Sample $B$ training samples $\{(\bx_i,y_i)\}_{i=1}^B$ from $\Ddistr_{tr}$.
\State Compute the raw gradient $\bg=- R(\{\bx_i,y_i\}_{i=1}^B)$.
\State Update moment estimates and compute the effective gradient $\bp$.
\State Update parameters:
 $\bTheta(t+1) = \bTheta(t) - \eta\bp - \eta\lambda\bTheta_{BN}(t)$ \textcolor{gray}{\# $\bTheta_{BN}$ is the batch normalization layers' parameters.}
\If{$t\%\tau==0$}
\State \textcolor{gray}{\# WRS update, assume $0-(L-1)$ layer is the convolution layer and the final layer is a fully connected layer. The convolution kernel has the form of $\bW_l\in\real^{K_l\times K_l\times C_{l,i}\times C_{l,o}}$ and the fully connected layer's weight has the form $\bW_l\in\real^{H_l\times K}$.}
\For{$l=0:L$}
\State \textcolor{gray}{\# Loop over the convolution layers}
\State \textcolor{gray}{\# Skip BN layers}
\State $\bW_{l}[:,:,:,i]=\bW_{l}[:,:,:,i]/\|\bW_{l}[:,:,:,i]\|_2\|,\forall i $.
\EndFor
\If{Classification Task}
\State $\bW_{L}[:,i]=\bW_{L}[:,i]/\|\bW_{L}[:,i]\|_2\|,\forall i$. \textcolor{gray}{\# Normalize the output layer for classification tasks.}
\EndIf
\EndIf
\EndWhile
\end{algorithmic}
\label{alg:wrs}
\end{algorithm}
\begin{table*}[h]
\centering 
\small
%\scriptsize
\newcolumntype{B}{ >{\centering\arraybackslash} m{3.0cm} }
\newcolumntype{C}{ >{\centering\arraybackslash} m{2.3cm} }
\newcolumntype{D}{ >{\centering\arraybackslash} m{3.5cm} }
\CUT{
\begin{tabular}{|D|B|C|C|C|}
\hline
Network+Opt. & Reg. & C10 & C100 & Tiny\\
\hline
  \multirow{3}{*}{VGG16+SGDM}&WD& 92.24(13) & 69.79(50) & 57.90(12)\\
  &WS+WD& \textbf{92.54(13)} & \textbf{70.84(21)} & 58.15(20)\\
&WRS& 92.37(17) & 69.69(28) & \textbf{58.28(04)}\\
\hline
  \multirow{3}{*}{VGG16+AdamW}&WD& 92.39(17) & 64.44(34) & 49.14(56)\\
  &WS+WD& 92.57(27) & 65.98(35) & 50.67(23)\\
&WRS& \textbf{93.16(21)} & \textbf{68.48(34)} & \textbf{52.15(73)}\\
\hline
  \multirow{3}{*}{ResNet18+SGDM}&WD& 94.37(04) & 76.13(34) & 62.58(20)\\
  & WS+WD & 94.38(04) & \textbf{76.78(31)} & 63.01(21) \\
&WRS& \textbf{94.41(07)} &76.77(09) & \textbf{64.02(17)}\\
\hline
  \multirow{3}{*}{ResNet18+AdamW}&WD& 93.44(11) & 72.83(11) & 57.34(13)\\
  & WS+WD & 93.28(11) & 73.84(10) & 58.76(18) \\
&WRS& \textbf{94.26(13)} & \textbf{75.95(15)} & \textbf{62.17(14)} \\
\hline
  \multirow{3}{*}{WideResNet+SGDM}&WD& 94.14(35) & \textbf{80.25(13)} & 66.82(16)\\
  & WS+WD & \textbf{95.47(03)} & 80.16(06) & 66.77(18) \\
&WRS& 95.22(15) & 79.95(18) & \textbf{69.22(08)} \\
\hline
  \multirow{3}{*}{WideResNet+AdamW}&WD& 94.17(15) & 74.97(15) & 60.57(20)\\
  & WS+WD & 92.46(16) & 75.34(18) & 61.70(35) \\
&WRS& \textbf{94.59(06)} & \textbf{78.89(05)} &  \textbf{64.03(26)}\\
\hline
\end{tabular}
}
\begin{tabular}{c@{\hspace{0.06cm}}|c@{\hspace{0.2cm}}c@{\hspace{0.2cm}}c|c@{\hspace{0.2cm}}c@{\hspace{0.2cm}}c|c@{\hspace{0.2cm}}c@{\hspace{0.2cm}}c}
\hline
Network+Opt. & \multicolumn{3}{c|}{CIFAR10} & \multicolumn{3}{c|}{CIFAR100} & \multicolumn{3}{c}{Tiny ImageNet}\\
& WD & WS & WRS & WD& WS& WRS & WD  & WS &WRS \\
\hline
  VGG16+SGDM
  & 92.24(.13)  %WD
  & \textbf{92.54(.13)} % WS+WD
  & 92.37(.17)   %WRS
  & 69.79(.50)  % WD
  & \textbf{70.84(.21)} % WS+WD
  & 69.69(.28)  % WRS
  & 57.90(.12)  % WD
   & 58.15(.20)  % WS+WD
   & \textbf{58.28(.04)} % WRS
   \\
  VGG16+Adam
  &92.39(.17)   % WD
  & 92.57(.27) %WS+WD
  & \textbf{93.16(.21)}   %WRS
  & 64.44(.34) % WD
  & 65.98(.35) %WS+WD
  & \textbf{68.48(.34)}  %WRS
  & 49.14(.56)% WD
  & 50.67(.23)%WS+WD
   & \textbf{52.15(.73)} %WRS
   \\
\hline
  ResNet18+SGDM
  &94.37(.04)   %WD
  & 94.38(.04) %WS+WD 
 & \textbf{94.41(.07)}  %WRS
  & 76.13(.34) %WD
  & \textbf{76.78(.31)} %WS+WD 
 &76.77(.09)   %WRS
  & 62.58(.20)%WD
  & 63.01(.21) %WS+WD   
 & \textbf{64.02(.17)} %WRS
 \\
 ResNet18+Adam
 & 93.44(.11) % WD 
 & 93.28(.11)  %WS+WD
& \textbf{94.26(.13)}  % WRS
 & 72.83(.11) % WD
 & 73.84(.10)  %WS+WD
  & \textbf{75.95(.15)} % WRS 
 & 57.34(.13)% WD
 & 58.76(.18) %WS+WD
 & \textbf{62.17(.14)} % WRS
  \\
\hline
  WideResNet+SGDM
 & 94.14(.35)   %WD
 & \textbf{95.47(.03)} % WS+WD
 & 95.22(.15)  % WRS
  & \textbf{80.25(.13)}  %WD
  & 80.16(.06) % WS+WD
  & 79.95(.18) % WRS
  & 66.82(.16) %WD
   & 66.77(.18) % WS+WD
   & \textbf{69.22(.08)} % WRS
  \\
 WideResNet+Adam
& 94.17(.15)  % WD
& 92.46(.16) % WS+WD
& \textbf{94.59(.06)}  %WRS
 & 74.97(.15)  % WD 
  & 75.34(.18)  % WS+WD 
  & \textbf{78.89(.05)}  %WRS 
 & 60.57(.20)  % WD
 & 61.70(.35)   % WS+WD
 &  \textbf{64.03(.26)} %WRS
  \\
\hline
\end{tabular}
\caption{Test accuracies of WD, WS (with WD) and WRS on image classification, averaged over 3 trials (standard deviation in parentheses). A paired t-test over all datasets, architectures, and optimizers indicates that WRS has overall higher accuracy (mean 76.9) versus WD (75.2) and WS (75.7); t(17)=4.287, p=0.0005 and t(17)=3.588, p=0.0023, respectively.
%The max value is 100, so samples could be 100,100,76 gives mean of 92, and stdev of ~11.3.  mabe you put the variance?}
}
\label{table:img_recog}
\end{table*}

\ziquan{In summary, we demonstrate that WRS has a similar regularization effect of inducing sparse conv kernels as WD, especially for early layers, while also having the benefit of stable training, robustness to hyperparameter value and achieves a competitive image classification performance. The next section provides a more systematic empirical study on the effectiveness of WRS. }

\section{Experiment}
\label{text:experiments}
In this section, we demonstrate the effectiveness of training BN-DNNs with weight rescaling (WRS), compared to weight decay (WD) and weight standardization with weight decay (WS+WD,) through  
%compared to BN+WD and BN+WS+WD through 
experiments on different neural networks and computer vision tasks.
\CUT{
\begin{figure}[t]
  \centering
  \includegraphics[width=0.5\linewidth]{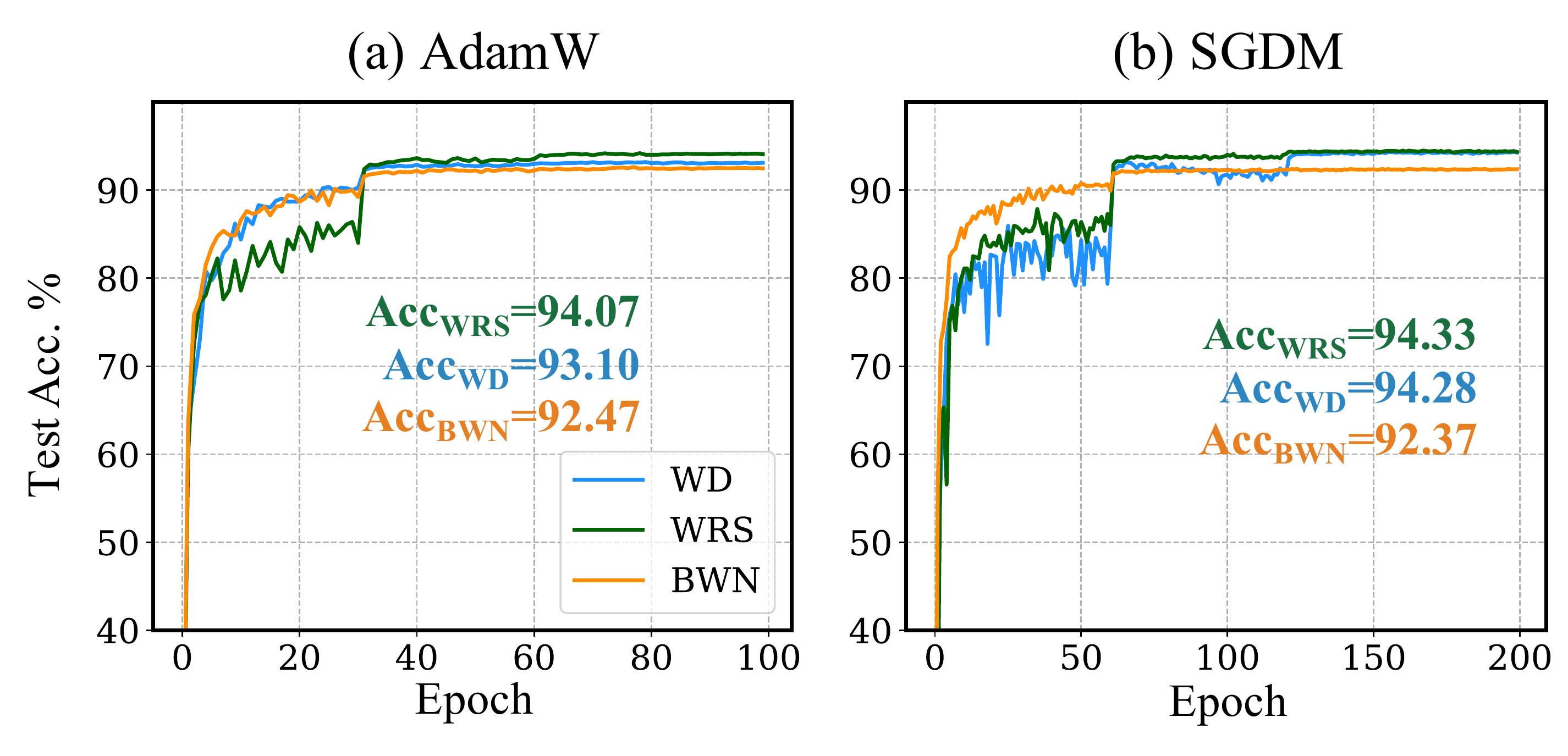}
  \caption{}
  \label{fig:BWN_WD_WRS}
\end{figure}
}
\subsection{Image Classification}
We apply WRS to different neural architectures and image classification datasets and compare the performance to WD, and WS+WD.

\subsubsection{Datasets}
 CIFAR10, CIFAR100 \cite{krizhevsky2009learning} and Tiny ImageNet \cite{deng2009imagenet} are used as the datasets. For CIFAR10 and CIFAR100, we use the same data preprocessing procedure: 1) normalize pixel values to $[0,1]$; 2) apply random horizontal flipping, random brightness, random contrast, random hue, random saturation and crop images after zero-padding 32$\times$32 images to 38$\times$38 ones. For Tiny ImageNet, we use the similar data preprocessing procedure but zero-pad 64$\times$64 images to 72$\times$72 before random cropping. 
 
 \subsubsection{Methods}
We test on three classical BN-DNNs, VGG16, ResNet18 \cite{he2016deep} and WideResNet-28-10 \cite{Zagoruyko2016WRN}. 
 Note that the original ResNet and WideResNet architectures have some conv layers without BN, and here we add BN for all convolution layers in this experiment to make a thorough comparison.\footnote{In Sec.~\ref{sec:adamp} and Sec.~\ref{sec:more_cv_task}, we show that WRS also achieves better results than the compared baselines for the standard ResNets, indicating that WRS is not affected by the reduced number of BN layers in ResNets. } Since SGD often has worse performance than SGD with momentum (SGDM) or Adam and is more frequently used in practice, we only use SGDM and Adam in our experiment. If AdamW is not explicitly mentioned, we use the original Adam in the experiment. For WD, we apply weight decay on all conv kernels and $\gamma$'s in the BN layers. For WRS, we  rescale conv kernels and weight vectors to the unit norm every $\tau$ optimization steps, and apply weight decay on parameters of BN layers. Note that the weight decay for $\gamma$ does not interfere with our analyses for \abc{fully-connected layer} weights in previous sections, since we do not make any assumptions on the dynamics of $\gamma$. In both WRS and WD, the hyperparameter for decay is searched from \{5e-3, 5e-4, 5e-5\} and batch size is 100. $\tau$ is chosen from $\{10,20,30,40,50\}$ based on their performance on the validation set. Finally, we do not test WeightNorm+WD since the performance of WN is shown to be generally worse than WD in image classification \cite{gitman2017comparison}.

%\TODO{need to reference the section numbers in the supplemental. also for other sentences referencing the supp.}.
\begin{figure}
  \centering
  \includegraphics[width=1.05\linewidth]{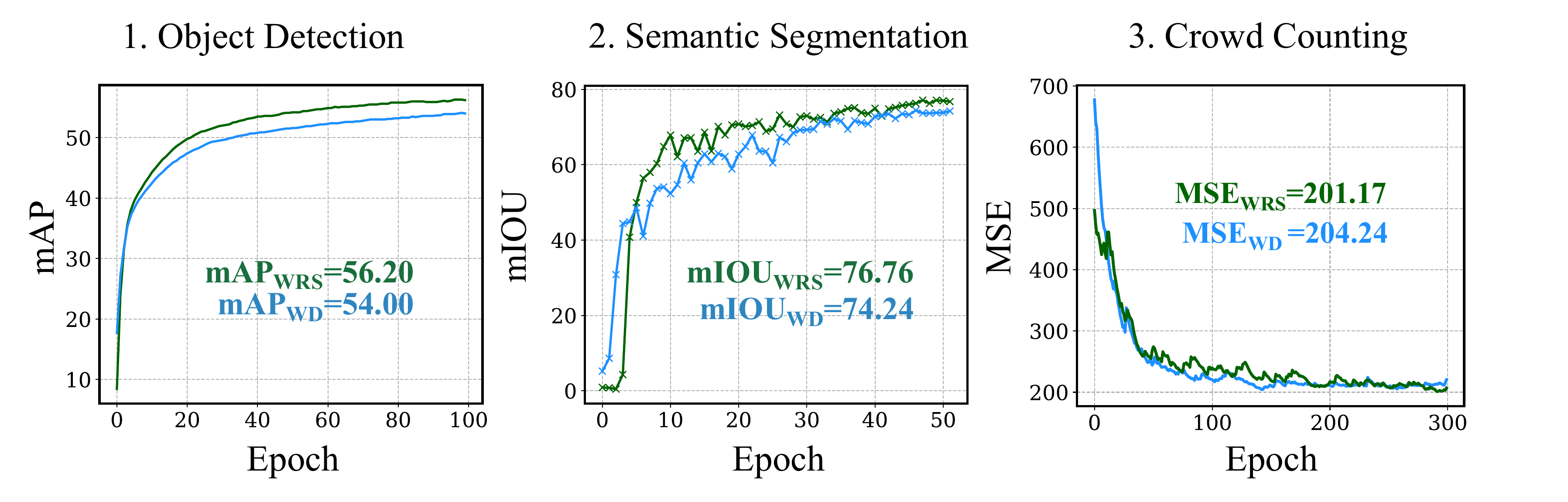}
  \caption{Learning curves of object detection and semantic segmentation. On the three CV tasks, WRS achieves faster convergence and improved generalization. }
  \label{fig:cv_tasks}
\end{figure}

\begin{figure*}[h]
  \centering
%  {\footnotesize \hspace{0.5cm} (a) with WRS \hspace{2.3cm} (b) with WD}
  \includegraphics[width=1.0\linewidth]{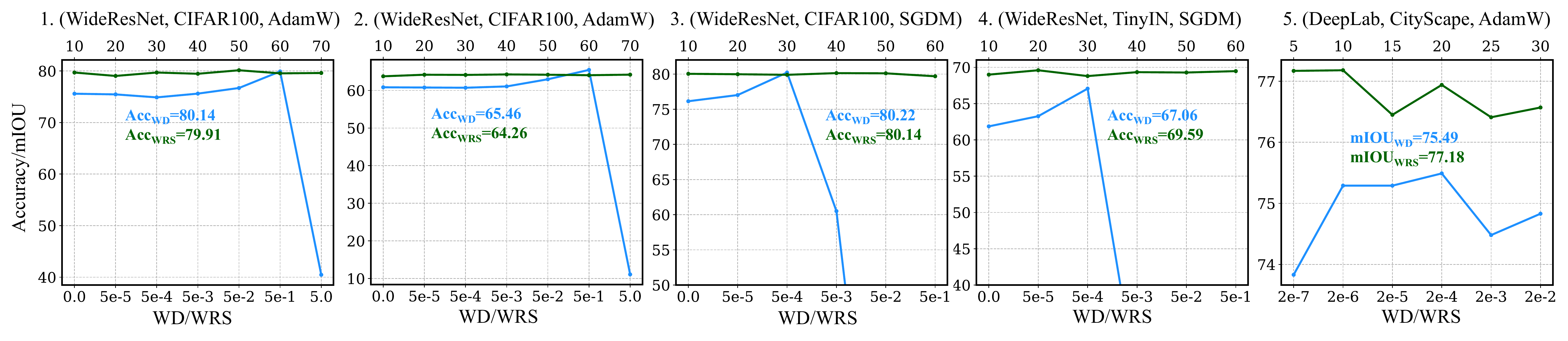}
  \caption{Effectiveness and robustness of WRS to hyperparameters. The best performance is denoted with colored text. On a variety of computer vision tasks, WRS has a comparable performance of WD and its robustness to hyperparameter values is stronger in SGDM and AdamW. On Tiny ImageNet and CityScape, WRS achieves better performance than WD.}
  \label{fig:robust_wrs}
\end{figure*}

\begin{figure}[t]
  \centering
  \includegraphics[width=1.0\linewidth]{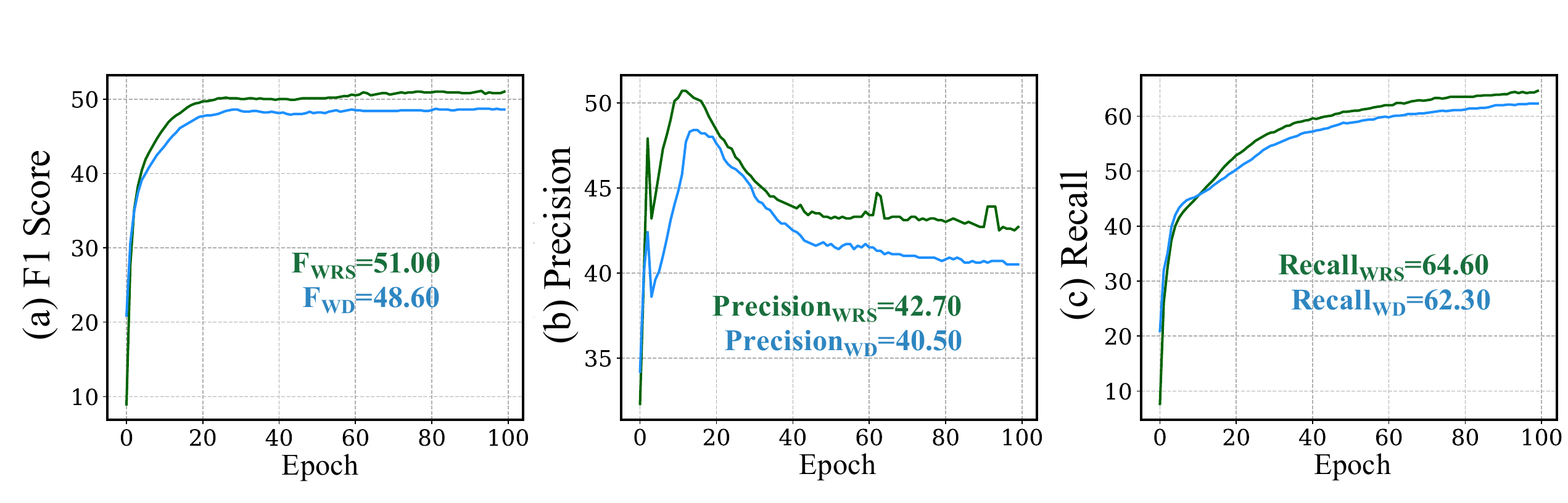}
  \caption{Object detection: comparison of a) F1 score, b) Precision, and c) Recall when training YOLOv3 with WRS and WD. }
  \label{fig:more_det_metric}
\end{figure}

\subsubsection{Results}
Table \ref{table:img_recog} shows a comparison of test accuracies for WRS, WD and WS+WD with various NNs, datasets, and optimization methods, where each accuracy is an average over running 3 trials. For most settings, especially for AdamW optimization, WRS outperforms WD in terms of test accuracy. Averaging over all datasets, architectures, and optimizers, WRS has overall higher accuracy (mean 76.9) versus WD (75.2) and WS (75.7), which is statistically significant via paired t-tests (see caption of Table \ref{table:img_recog}).
This result demonstrates that, in general, WRS improves generalization over WD when using advanced optimizers. Fig.~\ref{fig:robust_wrs} shows the result of WRS and WD using different hyperparameters when AdamW or SGDM is used in large-scale datasets. Although in some cases. WRS does not improve over WD, it is evident that the WRS has stronger robustness to hyperparameter selection than WD.
%Fig.~\ref{fig:cv_tasks}a shows the test  accuracy of training WideResNet on Tiny ImageNet with WRS and WD. WRS has a clear advantage over WD in this large-scale model and dataset scenario, in terms of stable training and better generalization. 

Fig.~\ref{fig:time_effect_and_resnet}b compares WRS with bounded WN (BWN) with mean-only BN \cite{hoffer2018norm} and WD. Although the convergence of BWN is faster than WRS and WD at the first training stage, BWN hurts the final accuracy in general. In addition to the performance benefit, WRS is easier to implement and robust to different NN architectures compared to BWN. For example, training with BWN on VGG16 always \ziquan{leads to exploding gradients and does not converge} in our experiment.

\subsection{Comparison of Variations of Adam}
\label{sec:adamp}
\ziquan{The differences between AdamP \cite{heo2021adamp} and WRS are discussed in Section \ref{text:related}. AdamP is proposed to increase the effective learning rate by projecting the momentum to the orthogonal space of a weight matrix, so that the update mainly alters the direction instead of radius of the vector. In contrast, WRS is a more straightforward way of constraining the weight norms by explicitly rescaling the weight. We compare the two optimization methods for training the standard ResNet18 and WideResNet on Tiny ImageNet.}

\ziquan{Table~\ref{tab:adamp_wrs} shows the result of AdamW, AdamP and their combination with WRS. We find that AdamP and AdamW+WRS are both effective in improving the performance of AdamW. In particular, on WideResNet, AdamW+WRS has a benefit over AdamP.}

\ziquan{In this experiment, we also try an input-channel normalization version of WRS, denoted as ``WRS(IC)'', where the rescaling operation divides $\bW[i,j,:,s]$ by $\|\bW[i,j,:,s]\|_2$, using the terminology in Algorithm~\ref{alg:wrs}. 
\abc{That is, with WRS(IC) the rescaling operation is applied separately on each spatial location of the convolution kernel.}
The WRS(IC) has a large benefit over WRS, especially on the larger network, since WRS(IC) also stablizes the weight norm and may encourage each kernel position to learn useful features. On Tiny ImageNet, WRS(IC) achieves the best result on both neural networks, indicating that the explicit rescaling is a promising technique to explore in the future.}
\begin{table}[]
    \centering
    \begin{tabular}{c|c|c}
    \hline
         Optimizer & Model & Test Acc. \\
         \hline
         \multirow{2}{*}{AdamW} & ResNet18 & 57.67(.25)\\
         & WideResNet & 60.98(.48)\\
         \hline
         \multirow{2}{*}{AdamP} & ResNet18 & 62.14(.36)\\
         & WideResNet & 62.91(.63)\\
         \hline
         \multirow{2}{*}{AdamW+WRS} & ResNet18 & 62.17(.46)\\
         & WideResNet & 63.30(.34)\\
         \hline
         \multirow{2}{*}{AdamP+WRS} & ResNet18 & 61.99(.39)\\
         & WideResNet & 63.05(.37)\\
         \hline
         \multirow{2}{*}{AdamW+WRS(IC)} & ResNet18 & 62.39(.25)\\
         & WideResNet & \textbf{66.27(.40)}\\
         \hline
         \multirow{2}{*}{AdamP+WRS(IC)} & ResNet18 & \textbf{62.80(.30)}\\
         & WideResNet & 65.96(.37)\\
         \hline
    \end{tabular}
    \caption{Comparison of AdamP and WRS on Tiny ImageNet with ResNet18 and WideResNet. Both AdamP and AdamW+WRS are effective with AdamW+WRS achieving the better result on WideResNet. The best performance for the two models is achieved by a variant of WRS that rescaling the conv kernel in the input channel dimension,  denoted as WRS(IC).}
    \label{tab:adamp_wrs}
\end{table}

\subsection{More Computer Vision Tasks}
\label{sec:more_cv_task}
We compare WRS with WD on BN-DNNs from more computer vision tasks. Different from the previous experiment where the model is trained from scratch, the training starts from a pre-trained model so the optimization is a fine-tuning process. The result shows that WRS has better performance than WD in transfer learning optimization.

\subsubsection{Object Detection} We use YOLOv3 \cite{redmon2018yolov3} as the model in the object detection task because of its heavy use of BatchNorm layers and faster training and inference. The original DarkNet53 \cite{redmon2013darknet} trained by ImageNet was chosen to initialize the backbone. We trained the network by 100 epochs with 0.01 initial learning rate and 8 batch size. The learning rate would change by the cosine schedule. The optimizer was SGDM with 0.0005 weight decay. The input size was set to 640$\times$320 and we use mAP@0.5 as the evaluation metrics. We use the large-scale COCO 2014 dataset \cite{lin2014microsoft}, which includes 82,783 training images, 40,504 validation images and 80 classes. The training is on 117,263 images from training and validation set, and the evaluation is on 5,000 validation images. 

Fig.~\ref{fig:cv_tasks}.1 shows that WRS is consistently better than WD after the first several epochs and converges faster than WD. The final mAP of WRS is 2.2 larger than that of WD. Fig.~\ref{fig:more_det_metric} shows the precision, recall and F score curve of WRS and WD during training, and WRS is better than WD in terms of all metrics.

\subsubsection{Semantic Segmentation} DeepLabv3 \cite{chen2017rethinking} is used in the semantic segmentation task since it has a BatchNorm layer after every convolution layer and these BatchNorm layers turn out to be essential to the training and generalization of DeepLabv3 \cite{chen2017rethinking}. We use ResNet101 as the backbone and initialize it with weights pre-trained on ImageNet and train DeepLabv3 using AdamW optimizer. Batch size is 10, initial learning rate is 1e-4, weight decay is 2e-4 and learning rate is scheduled as polynomial function where the end learning rate is 1e-6. The output stride is set as 16. The model is trained for 52 epochs. WRS uses a stepsize of 10. We use the augmented PASCAL VOC 2012 semantic segmentation dataset \cite{everingham2015pascal, hariharan2011semantic}. There are 10,582 training images and 1,449 validation images. We use the mean Intersection Over Union (mIOU) metric for semantic segmentation in this experiment. 

Fig.~\ref{fig:cv_tasks}.2 shows the mIOU %of DeepLabv3 
on the validation set during training. Our WRS achieves better performance (mIOU increase of 2.52) and faster convergence, compared to WD. Fig.~\ref{fig:robust_wrs}.5 shows the mIOU using different hyperparameters in WRS and WD, where WRS is always better than WD.

\subsubsection{Crowd Counting} We test the performance of WRS on this regression  task, where the goal is to predict a crowd density map from an input image. CSRNet \cite{li2018csrnet} is used as the model, where all convolution layers except for the output one are equipped with BN.  WRS is used for all conv layers before the output. VGG16 is selected as the backbone, which is initialized by an ImageNet pre-trained model. We minimize the Bayesian loss \cite{ma2019bayesian} in this experiment. UCF-QNRF \cite{idrees2018composition} is the dataset, containing 1201 training images and 334 test images. WD and WRS use the same weight decay parameter (1e-4) and learning rate (1e-4). The batch size is 32 and cropping size is 512. WRS uses a stepsize of 1. The model is trained for 300 epochs. No learning rate decay is used throughout the training. We split the training set into training and validation images and choose the model with the best MAE on the validation images as our final model. 

On the test set, WRS improves the MAE/MSE by 7.3/2.5 compared to WD (96.5/173.8 versus 103.8/176.3). Fig.~\ref{fig:cv_tasks}.3 shows the validation MSE during WRS and WD training, where WRS achieves better performance than WD at the end. To better visualize the trend, the curve is smoothed with an exponential moving average of 0.9. Note that in the QNRF dataset the validation set generally has more people than the test set so the MSE on the validation set is usually higher than that of the test set. 

\section{Conclusion}
In this paper, the weaknesses of WD in practical training are revealed, i.e. the strong sensitivity to hyperparameters and the overfitting effect after the learning rate decay. By reviewing the weight norm dynamics of BN-DNNs during training, we conclude that \abc{controlling weight norm} is necessary but WD is not the only choice. Therefore, we propose to regularize the training with the explicit weight rescaling (WRS) method. Our empirical studies demonstrate the effectiveness and robustness of WRS across various CV tasks, different neural architectures, and datasets. The limitation of this work is that the proposed WRS is only tested on major computer vision tasks so its benefit is unknown for other scale-invariant networks and applications like Transformer \cite{vaswani2017attention}. In future work, we will investigate the efficacy of WRS in other applications and scale-invariant neural architectures.

{\bf Acknowledgements.}
This work is supported by grants from the Research Grants Council of the Hong
Kong Special Administrative Region, China (CityU 11212518 and CityU 11215820).

{\appendix
\appendices

\section{Proof of Lemmas and Corollaries}
Here we provide proofs of the Lemmas and Corollaries in our paper.

\subsection{Proof of Lemma 3.1}
\noindent\textbf{Lemma 3.1} Suppose that we have a network as defined in the main paper, which is trained using SGD to minimize the loss $R_{\bTheta}$, then the gradient flow of $l_2$ norms of the weights in BatchNorm layers is zero, 
\begin{align}
\frac{d\|\bW_{l,j}(t)\|_2^2}{dt}=0.
\label{equ:thrm1}
\end{align}
\emph{Proof.} Using a minibatch of data $(\bX^{(B)},\bY^{(B)})=\{\bx_i,\by_i\}_{i=1}^B$, we derive the gradient of the empirical loss with respect to $\bW_{l,j}(t)$,
\begin{align}
\frac{\partial R_{\bTheta(t)}(\bX^{(B)},\bY^{(B)})}{\partial \bW_{l,j}(t)}=\sum_{i=1}^B\frac{\partial r^{(i)}(t)}{\partial \hat h_{l+1,j}^{(i)}(t)} \frac{\partial \hat h_{l+1,j}^{(i)}(t)}{\partial \bW_{l,j}(t)},\nonumber
\end{align}
where $\hat h_{l+1,j}^{i}(t)$ is the $i$th output of BatchNorm layer. The second gradient is
\begin{align}
\resizebox{0.5\textwidth}{!}{$
\frac{\partial \hat h_{l+1,j}^{(i)}}{\partial \bW_{l,j}}=\gamma_{l,j}\left[\frac{\bh_l^{(i)}-\bmu_l^{(B)}}{\|\bW_{l,j}\|_{\bSigma_{l}^{(B)}}}  - \frac{\bW_{l,j}^T(\bh_{l}^{(i)}-\bmu_{l}^{(B)})}{   \|\bW_{l,j}\|^3_{\bSigma_{l}^{(B)}}  } \bSigma_{l}^{(B)}\bW_{l,j}\right],$} \nonumber
\end{align}
where we omit $t$ for clarity. This gradient is equal to 
\begin{align}
\resizebox{0.5\textwidth}{!}{$\frac{\partial \hat h_{l+1,j}^{(i)}}{\partial \bW_{l,j}}=\gamma_{l,j}\left[\frac{\bh_l^{(i)}-\bmu_l^{(B)}}{(\bW_{l,j}^T\bSigma_{l}^{(B)}\bW_{l,j})^{1/2}}  -\frac{\bW_{l,j}^T(\bh_{l}^{(i)}-\bmu_{l}^{(B)})}{   (\bW_{l,j}^T\bSigma_{l}^{(B)}\bW_{l,j})^{3/2}  } \bSigma_{l}^{(B)}\bW_{l,j}\right].$}\nonumber
\end{align}
Consider the inner product of this gradient and $\bW_{l,j}$,
\begin{align*}
\resizebox{0.5\textwidth}{!}{$
\bW_{l,j}^T\frac{\partial \hat h_{l+1,j}^{(i)}}{\partial \bW_{l,j}}=\gamma_{l,j}\bW_{l,j}^T\left[\frac{\bh_l^{(i)}-\bmu_l^{(B)}}{(\bW_{l,j}^T\bSigma_{l}^{(B)}\bW_{l,j})^{1/2}}  - \frac{\bW_{l,j}^T(\bh_{l}^{(i)}-\bmu_{l}^{(B)})}{   (\bW_{l,j}^T\bSigma_{l}^{(B)}\bW_{l,j})^{3/2}  } \bSigma_{l}^{(B)}\bW_{l,j}\right]$}\\
\resizebox{0.5\textwidth}{!}{$=\gamma_{l,j}\left[\frac{\bW_{l,j}^T(\bh_l^{(i)}-\bmu_l^{(B)})}{(\bW_{l,j}^T\bSigma_{l}^{(B)}\bW_{l,j})^{1/2}}  - \frac{\bW_{l,j}^T(\bh_{l}^{(i)}-\bmu_{l}^{(B)})}{   (\bW_{l,j}^T\bSigma_{l}^{(B)}\bW_{l,j})^{3/2}  } \bW_{l,j}^T\bSigma_{l}^{(B)}\bW_{l,j}\right] $}\\
\resizebox{0.5\textwidth}{!}{$=\gamma_{l,j}\left[\frac{\bW_{l,j}^T(\bh_l^{(i)}-\bmu_l^{(B)})}{(\bW_{l,j}^T\bSigma_{l}^{(B)}\bW_{l,j})^{1/2}}  - \frac{\bW_{l,j}^T(\bh_{l}^{(i)}-\bmu_{l}^{(B)})}{   (\bW_{l,j}^T\bSigma_{l}^{(B)}\bW_{l,j})^{1/2}  }\right]=0.$}
\end{align*}
Similarly, if we consider the inner product of the gradient $\frac{\partial R_{\bTheta(t)}(\bX^{(B)},\bY^{(B)})}{\partial \bW_{l,j}(t)}$ and $\bW_{l,j}(t)$, we also have
\begin{align}
\Big\langle\frac{\partial R_{\bTheta(t)}(\bX^{(B)},\bY^{(B)})}{\partial \bW_{l,j}(t)},\bW_{l,j}(t)\Big\rangle =0.
\label{equ:ortho_grad}
\end{align}
Thus, the differential equation of $\bW_{l,j}(t)$ is
\begin{align*}
\frac{d\|\bW_{l,j}(t)\|_2^2}{dt}=-2 \Big\langle \frac{\partial R_{\bTheta(t)}(\bX^{(B)},\bY^{(B)})}{\partial \bW_{l,j}(t)},\bW_{l,j}(t) \Big\rangle=0.
\end{align*}
This complete the proof. Notice that we do not make any assumptions on the value of $\gamma_{l,j}$, so using weight decay for $\gamma_{l,j}$ does not invalidate this lemma. 

\subsection{Proof of Corollary 3.2}
\noindent\textbf{Corollary 3.2} Suppose that we have a network as in the main paper and the network is trained using gradient descent to minimize the loss $R_{\bTheta}$ and $l_2$ regularizations $\frac{\lambda}{2}\sum_{l=0}^{L}\|\bW_l\|_F^2$, then the weight $l_2$ norm in a BatchNorm layer follows an exponential decay,
\begin{align}
\|\bW_{l,j}(t)\|_2^2=\|\bW_{l,j}(0)\|_2^2\exp{(-2\lambda t)}.
\end{align}
\emph{Proof.} The ODE in SGD with weight decay is
\begin{align*}
\resizebox{0.5\textwidth}{!}{$
\frac{d\|\bW_{l,j}(t)\|_2^2}{dt}=\small -2 \Big\langle \frac{\partial R_{\bTheta(t)}(\bX^{(B)},\bY^{(B)})}{\partial \bW_{l,j}(t)}+\lambda\bW_{l,j}(t),\bW_{l,j}(t) \Big\rangle$}\\
\resizebox{0.5\textwidth}{!}{$=\small -2 \Big\langle \frac{\partial R_{\bTheta(t)}(\bX^{(B)},\bY^{(B)})}{\partial \bW_{l,j}(t)},\bW_{l,j}(t) \Big\rangle\small -2 \Big\langle \lambda\bW_{l,j}(t),\bW_{l,j}(t) \Big\rangle $}
\end{align*}
Using the result from Lemma 3.1 (\ref{equ:thrm1}), we know that the first term is 0. So if there is weight decay in the training, the dynamics of weight $l_2$ norm is
\begin{align}
\frac{d\|\bW_{l,j}(t)\|_2^2}{dt}=-2 \langle \lambda\bW_{l,j}(t),\bW_{l,j}(t) \rangle=-2\lambda\|\bW_{l,j}(t)\|_2^2.
\label{equ:wd_ode}
\end{align}
This ordinary differential equation has the form
\begin{align}
\dot x(t)=-2\lambda x(t), x(0)=x_0,
\end{align}
whose solution is $x(t)=x_0\exp(-2\lambda t)$. Similarly, the solution to the ODE in (\ref{equ:wd_ode}) is 
\begin{align}
\|\bW_{l,j}(t)\|_2^2=\|\bW_{l,j}(0)\|_2^2\exp{(-2\lambda t)}.
\label{eqn:discreteWt}
\end{align}
This completes the proof.
\subsection{Proof of Corollary 3.3}
\noindent\textbf{Corollary 3.3} Suppose that we have a network as in the main paper and the network is trained by SGD to minimize $R_{\bTheta}(\bX,\bY)$, then we have
\begin{align}
\bDelta_{l,j}(t)=\eta(t)^2\|\frac{\partial R_{\bTheta(t)}(\bX^{(B)},\bY^{(B)})}{\partial \bW_{l,j}(t)}\|_2^2.
\end{align}
\emph{Proof.} We derive the corollary as follows
\begin{align*}
\small
\bDelta_{l,j}(t)=&\|\bW_{l,j}(t+1)\|_2^2-\|\bW_{l,j}(t)\|_2^2\\
=&\small\|\bW_{l,j}(t)-\eta(t)\frac{\partial R_{\bTheta(t)}(\bX^{(B)},\bY^{(B)})}{\partial \bW_{l,j}(t)}\|_2^2-\|\bW_{l,j}(t)\|_2^2\\
=&\small\|\bW_{l,j}(t)\|_2^2-2\eta(t) \Big\langle\bW_{l,j}(t),\frac{\partial R_{\bTheta(t)}(\bX^{(B)},\bY^{(B)})}{\partial \bW_{l,j}(t)} \Big\rangle +\\
&\small\eta(t)^2\|\frac{\partial R_{\bTheta(t)}(\bX^{(B)},\bY^{(B)})}{\partial \bW_{l,j}(t)}\|_2^2-\|\bW_{l,j}(t)\|_2^2.
\end{align*}
Note that in Lemma 3.1, we do not use the continuous-time property when deriving (\ref{equ:ortho_grad}), and thus (\ref{equ:ortho_grad}) still holds in the discrete-time domain. Thus, we have $\big\langle\bW_{l,j}(t),\frac{\partial R_{\bTheta(t)}(\bX^{(B)},\bY^{(B)})}{\partial \bW_{l,j}(t)}\big\rangle=0$, and substituting yields (\ref{eqn:discreteWt}).

\section{Formulation of Convolution Neural Networks}
As we explain in the main paper, convolution neural networks (CNNs) are special cases of fully-connected neural networks. Now we show that the lemmas and corollaries still hold for CNNs with BatchNorm. A weight in a MLP $\bW_{l}\in\real^{H_l\times H_{l+1}}$ corresponds to a convolution kernel $\bW_{l}^{(\mathcal{K})}\in\real^{K_l\times K_l \times C_{l,i} \times C_{l,o}}$ in a CNN, where $K_l$ is the size of conv kernel, $C_{l,i}$ and $C_{l,o}$ are the number of input and output channels. The BatchNorm for one neuron in a MLP corresponds to the BatchNorm for one conv channel in a CNN. If we use $\bW_{l,j}^{(\mathcal{K})}\in\real^{K_l\times K_l \times C_{l,i} \times 1}, j=1,\cdots,C_{l,o}$ to denote the conv kernel for $j$th output channel and replace $\bW_{l,j}$ with $\bW_{l,j}^{(\mathcal{K})}$ in previous sections, our derivations also holds. The only difference is that a conv kernel is applied multiple times while a weight is multiplied once for one input point. So some summation terms need to be rewritten as a sum over input image patches. For example, the gradient of empirical risk with respect to the conv kernel $\bW_{l,j}^{(\mathcal{K})}$ is 
\begin{align*}
\resizebox{0.5\textwidth}{!}{$
\frac{\partial R_{\bTheta(t)}(\bX^{(B)},\bY^{(B)})}{\partial \bW_{l,j}^{(\mathcal{K})}(t)}=\sum_{i=1}^B\sum_{m=1}^{S_{l+1}}\frac{\partial r^{(i)}(t)}{\partial \hat h_{l+1,m,j}^{(i)}(t)} \frac{\partial \hat h_{l+1,m,j}^{(i)}(t)}{\partial \bW_{l,j}^{(\mathcal{K})}(t)},$}
\end{align*}
where $S_{l+1}$ is the number features in one channel of $(l+1)$th layer. The second gradient is
\begin{align*}
\resizebox{0.5\textwidth}{!}{$
\frac{\partial \hat h_{l+1,m, j}^{(i)}}{\partial \bW_{l,j}^{(\mathcal{K})}}=\gamma_{l,j}\left[\frac{\bh_l^{(i)}-\bmu_l^{(B)}}{((\bW_{l,j}^{(\mathcal{K})})^T\bSigma_{l}^{(B)}\bW_{l,j}^{(\mathcal{K})})^{1/2}}-\frac{{\bW_{l,j}^{(\mathcal{K})}}^T(\bh_{l}^{(i)}-\bmu_{l}^{(B)})}{   ((\bW_{l,j}^{(\mathcal{K})})^T\bSigma_{l}^{(B)}\bW_{l,j}^{(\mathcal{K})})^{3/2}} \bSigma_{l}^{(B)}\bW_{l,j}^{(\mathcal{K})}\right],$}
\end{align*}
where $\bh_l^{(i)}$ is the image/feature patch for its output $\hat h_{l+1,m, j}^{(i)}$, $\bmu_l^{(B)}$ and $\bSigma_{l}^{(B)}$ are the mean and covariance computed on image/feature patches for the convolution. We can multiply $\bW_{l,j}^{(\mathcal{K})}$ with the gradient and get the same result as Lemma 3.1 for CNNs. Corollary 3.2 and 3.3 can also be easily derived for CNNs. 
}
{\small
\bibliographystyle{ieee_fullname}
\bibliography{ref}

\begin{thebibliography}{10}\itemsep=-1pt

\bibitem{arora2018optimization}
Sanjeev Arora, Nadav Cohen, and Elad Hazan.
\newblock On the optimization of deep networks: Implicit acceleration by
  overparameterization.
\newblock In {\em International Conference on Machine Learning}, pages
  244--253, 2018.

\bibitem{arora2018theoretical}
Sanjeev Arora, Zhiyuan Li, and Kaifeng Lyu.
\newblock Theoretical analysis of auto rate-tuning by batch normalization.
\newblock {\em arXiv preprint arXiv:1812.03981}, 2018.

\bibitem{chen2017rethinking}
Liang-Chieh Chen, George Papandreou, Florian Schroff, and Hartwig Adam.
\newblock Rethinking atrous convolution for semantic image segmentation.
\newblock {\em arXiv preprint arXiv:1706.05587}, 2017.

\bibitem{cho2017riemannian}
Minhyung Cho and Jaehyung Lee.
\newblock Riemannian approach to batch normalization.
\newblock In I. Guyon, U.~Von Luxburg, S. Bengio, H. Wallach, R. Fergus, S.
  Vishwanathan, and R. Garnett, editors, {\em Advances in Neural Information
  Processing Systems}, volume~30. Curran Associates, Inc., 2017.

\bibitem{deng2009imagenet}
Jia Deng, Wei Dong, Richard Socher, Li-Jia Li, Kai Li, and Li Fei-Fei.
\newblock Imagenet: A large-scale hierarchical image database.
\newblock In {\em 2009 IEEE conference on computer vision and pattern
  recognition}, pages 248--255. Ieee, 2009.

\bibitem{du2018gradient}
Simon~S Du, Xiyu Zhai, Barnabas Poczos, and Aarti Singh.
\newblock Gradient descent provably optimizes over-parameterized neural
  networks.
\newblock In {\em International Conference on Learning Representations}, 2018.

\bibitem{everingham2015pascal}
Mark Everingham, SM~Ali Eslami, Luc Van~Gool, Christopher~KI Williams, John
  Winn, and Andrew Zisserman.
\newblock The {PASCAL} visual object classes challenge: A retrospective.
\newblock {\em International journal of computer vision}, 111(1):98--136, 2015.

\bibitem{gitman2017comparison}
Igor Gitman and Boris Ginsburg.
\newblock Comparison of batch normalization and weight normalization algorithms
  for the large-scale image classification.
\newblock {\em arXiv preprint arXiv:1709.08145}, 2017.

\bibitem{golatkar2019time}
Aditya~Sharad Golatkar, Alessandro Achille, and Stefano Soatto.
\newblock Time matters in regularizing deep networks: Weight decay and data
  augmentation affect early learning dynamics, matter little near convergence.
\newblock In {\em Advances in Neural Information Processing Systems}, pages
  10678--10688, 2019.

\bibitem{hariharan2011semantic}
Bharath Hariharan, Pablo Arbel{\'a}ez, Lubomir Bourdev, Subhransu Maji, and
  Jitendra Malik.
\newblock Semantic contours from inverse detectors.
\newblock In {\em 2011 International Conference on Computer Vision}, pages
  991--998. IEEE, 2011.

\bibitem{he2016deep}
Kaiming He, Xiangyu Zhang, Shaoqing Ren, and Jian Sun.
\newblock Deep residual learning for image recognition.
\newblock In {\em Proceedings of the IEEE conference on computer vision and
  pattern recognition}, pages 770--778, 2016.

\bibitem{heo2021adamp}
Byeongho Heo, Sanghyuk Chun, Seong~Joon Oh, Dongyoon Han, Sangdoo Yun, Gyuwan
  Kim, Youngjung Uh, and Jung-Woo Ha.
\newblock Adamp: Slowing down the slowdown for momentum optimizers on
  scale-invariant weights.
\newblock In {\em International Conference on Learning Representations}, 2021.

\bibitem{hoffer2018norm}
Elad Hoffer, Ron Banner, Itay Golan, and Daniel Soudry.
\newblock Norm matters: efficient and accurate normalization schemes in deep
  networks.
\newblock In {\em Advances in Neural Information Processing Systems}, pages
  2160--2170, 2018.

\bibitem{huang2017projection}
Lei Huang, Xianglong Liu, Bo Lang, and Bo Li.
\newblock Projection based weight normalization for deep neural networks.
\newblock {\em arXiv preprint arXiv:1710.02338}, 2017.

\bibitem{huang2017centered}
Lei Huang, Xianglong Liu, Yang Liu, Bo Lang, and Dacheng Tao.
\newblock Centered weight normalization in accelerating training of deep neural
  networks.
\newblock In {\em Proceedings of the IEEE International Conference on Computer
  Vision}, pages 2803--2811, 2017.

\bibitem{idrees2018composition}
Haroon Idrees, Muhmmad Tayyab, Kishan Athrey, Dong Zhang, Somaya Al-Maadeed,
  Nasir Rajpoot, and Mubarak Shah.
\newblock Composition loss for counting, density map estimation and
  localization in dense crowds.
\newblock In {\em Proceedings of the European Conference on Computer Vision
  (ECCV)}, pages 532--546, 2018.

\bibitem{ioffe2015batch}
Sergey Ioffe and Christian Szegedy.
\newblock Batch normalization: Accelerating deep network training by reducing
  internal covariate shift.
\newblock {\em arXiv preprint arXiv:1502.03167}, 2015.

\bibitem{kingma2014adam}
Diederik~P Kingma and Jimmy Ba.
\newblock Adam: A method for stochastic optimization.
\newblock {\em arXiv preprint arXiv:1412.6980}, 2014.

\bibitem{krizhevsky2009learning}
Alex Krizhevsky et~al.
\newblock Learning multiple layers of features from tiny images.
\newblock 2009.

\bibitem{li2020understanding}
Xiang Li, Shuo Chen, and Jian Yang.
\newblock Understanding the disharmony between weight normalization family and
  weight decay.
\newblock In {\em AAAI}, pages 4715--4722, 2020.

\bibitem{li2019towards}
Yuanzhi Li, Colin Wei, and Tengyu Ma.
\newblock Towards explaining the regularization effect of initial large
  learning rate in training neural networks.
\newblock In {\em Advances in Neural Information Processing Systems}, pages
  11674--11685, 2019.

\bibitem{li2018csrnet}
Yuhong Li, Xiaofan Zhang, and Deming Chen.
\newblock {CSR}net: Dilated convolutional neural networks for understanding the
  highly congested scenes.
\newblock In {\em Proceedings of the IEEE conference on computer vision and
  pattern recognition}, pages 1091--1100, 2018.

\bibitem{li2019exponential}
Zhiyuan Li and Sanjeev Arora.
\newblock An exponential learning rate schedule for deep learning.
\newblock {\em arXiv preprint arXiv:1910.07454}, 2019.

\bibitem{lin2014microsoft}
Tsung-Yi Lin, Michael Maire, Serge Belongie, James Hays, Pietro Perona, Deva
  Ramanan, Piotr Doll{\'a}r, and C~Lawrence Zitnick.
\newblock Microsoft {COCO}: Common objects in context.
\newblock In {\em European conference on computer vision}, pages 740--755.
  Springer, 2014.

\bibitem{liu2020improve}
Ziquan Liu, Yufei Cui, and Antoni~B Chan.
\newblock Improve generalization and robustness of neural networks via weight
  scale shifting invariant regularizations.
\newblock {\em arXiv preprint arXiv:2008.02965}, 2020.

\bibitem{loshchilov2018decoupled}
Ilya Loshchilov and Frank Hutter.
\newblock Decoupled weight decay regularization.
\newblock In {\em International Conference on Learning Representations}, 2018.

\bibitem{ma2019bayesian}
Zhiheng Ma, Xing Wei, Xiaopeng Hong, and Yihong Gong.
\newblock Bayesian loss for crowd count estimation with point supervision.
\newblock In {\em Proceedings of the IEEE International Conference on Computer
  Vision}, pages 6142--6151, 2019.

\bibitem{nadarajah2005generalized}
Saralees Nadarajah.
\newblock A generalized normal distribution.
\newblock {\em Journal of Applied statistics}, 32(7):685--694, 2005.

\bibitem{neyshabur2015path}
Behnam Neyshabur, Russ~R Salakhutdinov, and Nati Srebro.
\newblock Path-{SGD}: Path-normalized optimization in deep neural networks.
\newblock In {\em Advances in Neural Information Processing Systems}, pages
  2422--2430, 2015.

\bibitem{qiao2019weight}
Siyuan Qiao, Huiyu Wang, Chenxi Liu, Wei Shen, and Alan Yuille.
\newblock Weight standardization.
\newblock {\em arXiv preprint arXiv:1903.10520}, 2019.

\bibitem{redmon2013darknet}
Joseph Redmon.
\newblock Darknet: Open source neural networks in c, 2013.

\bibitem{redmon2018yolov3}
Joseph Redmon and Ali Farhadi.
\newblock {YOLO}v3: An incremental improvement.
\newblock {\em arXiv preprint arXiv:1804.02767}, 2018.

\bibitem{Roburin2020ASA}
Simon Roburin, Yann de Mont-Marin, Andrei Bursuc, Renaud Marlet, Patrick
  P'erez, and Mathieu Aubry.
\newblock A spherical analysis of adam with batch normalization.
\newblock {\em arXiv: Learning}, 2020.

\bibitem{salimans2016weight}
Tim Salimans and Durk~P Kingma.
\newblock Weight normalization: A simple reparameterization to accelerate
  training of deep neural networks.
\newblock In {\em Advances in neural information processing systems}, pages
  901--909, 2016.

\bibitem{simonyan2014very}
Karen Simonyan and Andrew Zisserman.
\newblock Very deep convolutional networks for large-scale image recognition.
\newblock {\em arXiv preprint arXiv:1409.1556}, 2014.

\bibitem{van2017l2}
Twan Van~Laarhoven.
\newblock L2 regularization versus batch and weight normalization.
\newblock {\em arXiv preprint arXiv:1706.05350}, 2017.

\bibitem{vaswani2017attention}
Ashish Vaswani, Noam Shazeer, Niki Parmar, Jakob Uszkoreit, Llion Jones,
  Aidan~N Gomez, {\L}ukasz Kaiser, and Illia Polosukhin.
\newblock Attention is all you need.
\newblock In {\em Advances in neural information processing systems}, pages
  5998--6008, 2017.

\bibitem{Zagoruyko2016WRN}
Sergey Zagoruyko and Nikos Komodakis.
\newblock Wide residual networks.
\newblock In {\em BMVC}, 2016.

\bibitem{zhang2020spherical}
Dingyi Zhang, Yingming Li, and Zhongfei Zhang.
\newblock Deep metric learning with spherical embedding.
\newblock In H. Larochelle, M. Ranzato, R. Hadsell, M.F. Balcan, and H. Lin,
  editors, {\em Advances in Neural Information Processing Systems}, volume~33,
  pages 18772--18783. Curran Associates, Inc., 2020.

\bibitem{zhang2018three}
Guodong Zhang, Chaoqi Wang, Bowen Xu, and Roger Grosse.
\newblock Three mechanisms of weight decay regularization.
\newblock In {\em International Conference on Learning Representations}, 2018.

\end{thebibliography}
}

\vfill

\end{document}